\documentclass{article}
\usepackage{iclr2026_conference,times}

\usepackage{amsmath,amsfonts,bm}

\def\eqref#1{equation~\ref{#1}}
\def\1{\bm{1}}

\DeclareMathAlphabet{\mathsfit}{\encodingdefault}{\sfdefault}{m}{sl}
\SetMathAlphabet{\mathsfit}{bold}{\encodingdefault}{\sfdefault}{bx}{n}

\usepackage{hyperref}
\usepackage{url}
\usepackage{latexsym}
\usepackage{adjustbox}
\usepackage{xspace}
\usepackage{todonotes}
\usepackage[normalem]{ulem}
\usepackage{comment}
\usepackage{tikz}
\usepackage{makecell}
\usepackage{adjustbox}
\usepackage{pifont}
\usepackage{float}
\usepackage{supertabular}
\usepackage{multirow}
\usepackage{longtable}
\usepackage{graphicx}
\usepackage[edges]{forest}
\definecolor{hidden-draw}{RGB}{0,0,0}
\usepackage{wrapfig}
\usepackage{CJKutf8}
\usepackage{CJK}
\usepackage{subcaption}

\usepackage{tcolorbox}
\tcbset{
    mybox/.style={
        fontupper=\linespread{.8}\selectfont,
        sharp corners,
        top=0pt, % 设置顶部留白大小
        bottom=0pt % 设置底部留白大小
    }
}

\usepackage{amsmath}
\usepackage{amssymb}

\usepackage{booktabs}
\usepackage{threeparttable}
\usepackage{tabularx}

\usepackage{soul}

\usepackage{multirow}

\usepackage{bm} % bold math

\usepackage{algorithm}
\usepackage{algorithmic}

\usepackage{setspace}

\newcommand{\ctext}[2]{%
    \tikz[baseline=(X.base)] \node[fill=#1,rounded corners=2.8pt,inner sep=1pt] (X) {#2};%
}

\definecolor{mygreen}{RGB}{11,141,10}
\definecolor{myred}{RGB}{223,68,52}
\definecolor{myblue}{RGB}{70,130,180}
\definecolor{mydeepblue}{RGB}{65,105,225}
\definecolor{myviolet}{RGB}{97,0,138}
\definecolor{myburgundy}{RGB}{110,10,30}
\definecolor{myblue2}{RGB}{0,105,148}
\definecolor{iceblue}{RGB}{173, 216, 230}
\definecolor{puregreen}{RGB}{0, 218, 0}
\definecolor{graygreen}{RGB}{74,113,106}
\definecolor{darkblue}{rgb}{0.0,0.0,0.55}
\definecolor{wingreen}{rgb}{0,0.45,0.24}
\definecolor{losered}{rgb}{1.0,0.1,0.24}

\definecolor{lightcoral}{rgb}{0.97, 0.36, 0.46}
\definecolor{lightyellow}{rgb}{0.98, 0.7, 0}
\definecolor{harvestgold}{rgb}{0.85, 0.57, 0.0}
\definecolor{brightlavender}{rgb}{0.75, 0.58, 0.89}
\definecolor{capri}{rgb}{0.0, 0.75, 1.0}
\definecolor{carminepink}{rgb}{0.92, 0.3, 0.26}
\definecolor{celadon}{rgb}{0.67, 0.88, 0.69}
\definecolor{darkpastelgreen}{rgb}{0.01, 0.75, 0.24}

\definecolor{grayhighlight}{RGB}{250,250,227}

\definecolor{target}{HTML}{F47983}
\definecolor{control}{HTML}{3E87CD}
\definecolor{credibility}{HTML}{B98AC9}
\definecolor{logical}{HTML}{93C572}
\definecolor{emotional}{HTML}{F9EAC3}

\newenvironment{packeditemize}{
\begin{list}{$\bullet$}{
\setlength{\labelwidth}{8pt}
\setlength{\itemsep}{0pt}
\setlength{\leftmargin}{\labelwidth}
\addtolength{\leftmargin}{\labelsep}
\setlength{\parindent}{0pt}
\setlength{\listparindent}{\parindent}
\setlength{\parsep}{0pt}
\setlength{\topsep}{3pt}}}{\end{list}}

\usepackage{pifont}

\newcommand{\diffdown}[1]{\raisebox{0.5pt}{\fontsize{6}{5.5}\selectfont{\textcolor{wingreen}{\textbf{$\blacktriangledown${$ #1$}}}}}}
\newcommand{\diffup}[1]{\raisebox{0.5pt}{\fontsize{6}{5.5}\selectfont{\textcolor{losered}{\textbf{$\blacktriangle${$ #1$}}}}}}

\newcommand{\placeholder}[1]{\ctext{yellow!40}{#1}}

\definecolor{orange}{HTML}{E66100}

\newcommand{\myline}{\par
  \kern0pt 
  \hrule height 0.6pt
  \kern3pt 
}

\newcommand{\mylinenoskip}{\par
  \kern3pt 
  \hrule height 0.6pt
  \kern3pt 
}
\newtcbox{\hlsecondarytab}{on line, box align=base, ,colback=orange!20!white,colframe=orange,size=fbox,arc=3pt, before upper=\strut, top=-2.5pt, bottom=-4.5pt, left=1pt, right=1pt, boxrule=0pt}

\newcommand{\sys}{\textsc{unPact}\xspace}
\newcommand{\key}{\textsc{KeyTokens}\xspace}
\newcommand{\recover}{\textsc{FocusOnKey}\xspace}
\newcommand{\baseline}{\textsc{Probab}\xspace}

\newcommand{\GA}{\textsf{GA}\xspace}
\newcommand{\GDR}{\textsf{GD}\xspace}
\newcommand{\KLR}{\textsf{KL}\xspace}
\newcommand{\GAGDR}{$\textsf{GA}_\GDR$\xspace}
\newcommand{\GAKLR}{$\textsf{GA}_\KLR$\xspace}
\newcommand{\NPO}{\textsf{NPO}\xspace}
\newcommand{\NPOGDR}{$\textsf{NPO}_\GDR$\xspace}
\newcommand{\NPOKLR}{$\textsf{NPO}_\KLR$\xspace}
\newcommand{\AVERAGE}{\textsf{Average}\xspace}
\newcommand{\TV}{\textsf{TV}\xspace}

\newcommand{\RMU}{\textsf{RMU}\xspace}

\usepackage[table,xcdraw]{xcolor}
\usepackage{colortbl}
\usepackage[normalem]{ulem} % 提供下划线功能

\definecolor{darkred}{RGB}{139,0,0}
\newcommand{\mycolorbox}[2][red!30]{%
  {\setlength{\fboxsep}{1pt}\colorbox{#1}{#2}}%
}
\definecolor{scired}{RGB}{254, 224, 210}
\definecolor{sciblue}{RGB}{222, 235, 247}

\newtcbox{\hlyellow}{on line, box align=base, colback=scired,colframe=white,size=fbox,arc=2pt, before upper=\strut, top=-3pt, bottom=-4.5pt, left=-2pt, right=-2pt, boxrule=0pt}

\newtcbox{\hlwhite}{on line, box align=base, colback=sciblue,colframe=white,size=fbox,arc=2pt, before upper=\strut, top=-3pt, bottom=-4.5pt, left=-2pt, right=-2pt, boxrule=0pt}

\newcommand{\reduline}[1]{\textcolor{red}{\uline{\textcolor{black}{#1}}}}

\title{Understanding the Dilemma of Unlearning for Large Language Models}

\author{
    \centerline{Qingjie Zhang\textsuperscript{1}, \ 
    Haoting Qian\textsuperscript{1}, \ 
    Zhicong Huang\textsuperscript{2}, \ 
    Cheng Hong\textsuperscript{2},
    } \vspace{0.5mm}  \\
    \centerline{\textbf{
    Minlie Huang\textsuperscript{1}, \
    Ke Xu\textsuperscript{1}, \ 
    Chao Zhang\textsuperscript{1}, \ 
    and Han Qiu\textsuperscript{1}\thanks{The corresponding author}
    }} \vspace{0.5mm} \\
    \centerline{\normalsize{$^{1}$Tsinghua University, $^{2}$Ant Group}} \vspace{0.5mm} \\
    \centerline{\texttt{Emails:\small \{qj-zhang24@mails., qiuhan@\}tsinghua.edu.cn}}
}

\iclrfinalcopy 
\begin{document}

\maketitle

\begin{abstract}
Unlearning seeks to remove specific knowledge from large language models (LLMs), but its effectiveness remains contested. On one side, ``forgotten'' knowledge can often be recovered through interventions such as light fine-tuning; on the other side, unlearning may induce catastrophic forgetting that degrades general capabilities. Despite active exploration of unlearning methods, interpretability analyses of the mechanism are scarce due to the difficulty of tracing knowledge in LLMs’ complex architectures.
We address this gap by proposing \sys, an interpretable framework for unlearning via prompt attribution and contribution tracking. Typically, \emph{it quantifies each prompt token's influence on outputs, enabling pre- and post-unlearning comparisons to reveal what changes.} Across six mainstream unlearning methods, three LLMs, and three benchmarks, we find that: \\
(1) Unlearning appears to be effective by disrupting focus on keywords in prompt; \\
(2) Much of the knowledge is not truly erased and can be recovered by simply emphasizing these keywords in prompts, without modifying the model’s weights;\\
(3) Catastrophic forgetting arises from indiscriminate penalization of all tokens.
Together, our results suggest an unlearning dilemma: existing methods tend either to be insufficient - knowledge remains recoverable by keyword emphasis, or destructive - general performance collapses due to catastrophic forgetting, leaving a gap to reliable unlearning.
We open-source at \url{https://unpact.site}.
\end{abstract}

\section{Introduction}

\begin{wrapfigure}{r}{0.33\linewidth}
    \vspace{-3ex}
    \centering
    \includegraphics[width=0.99\linewidth]{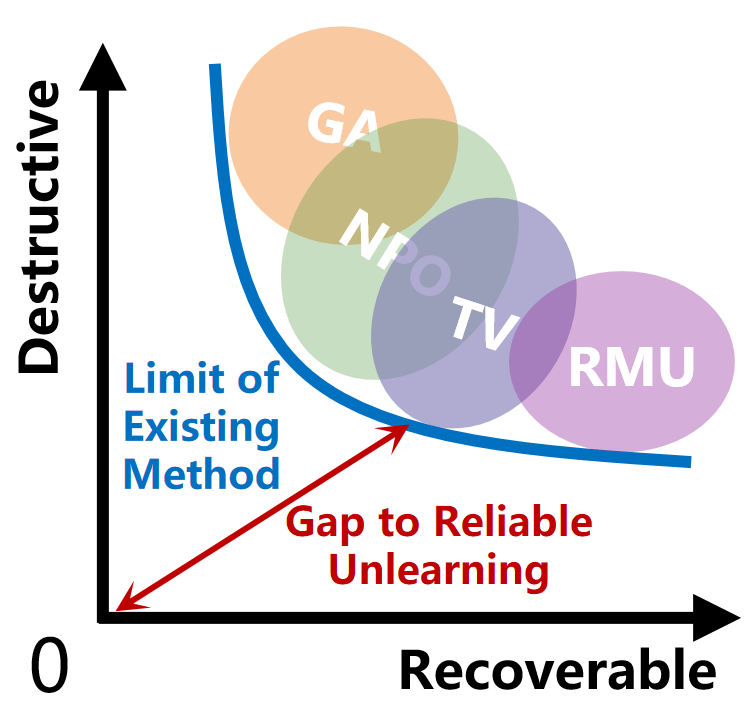}
    \vspace{-2.5ex}
    \caption{Dilemma of unlearning: either recoverable by keyword emphasis, or destructive to catastrophic forgetting.
    }
    \label{fig:dilemma}
    \vspace{-2.7ex}
\end{wrapfigure}
Unlearning aims to remove or suppress specific data or knowledge from a trained model \citep{thaker2025position,jia2024soul,yuan2024closer}. Yet the effectiveness on large language models (LLMs) remains contested, presenting a dilemma: on one side, purportedly ``forgotten'' knowledge can be recovered by sophisticated interventions such as model weight editing \citep{patil2023can} or additional fine-tuning \citep{hu2025unlearning,lynch2024eight}; on the other side, aggressive unlearning may induce catastrophic forgetting, degrading performance on capabilities beyond the intended targets \citep{li2024revisiting,luo2023empirical}. This tension motivates a deeper understanding of how unlearning actually operates in LLMs.

Despite active progress on unlearning algorithms, there is a shortage of interpretability analyses. The primary challenge arises from the inherent difficulty of tracking the specific knowledge during unlearning within LLMs' complex architecture \citep{hakimi2025time,wang2025tracing,li2024knowledge}. For open-sourced LLMs where internal weights are accessible, interpreting how knowledge is stored and removed is hard due to intricate weight interactions \citep{zhao2024explainability}. And for closed-source LLMs, interpretability is harder as it must rely solely on inputs and outputs of models, limiting access to internals that many classic interpretability tools require \citep{mumuni2025explainable,zhao2024explainability,dwivedi2023explainable}.

Building on this limitation, we address the gap by interpreting unlearning from the prompt perspective, which directly affects the output and is agnostic to whether the LLM is open- or closed-source. We first propose \sys, an interpretable framework for \ul{un}learning through \ul{p}rompt \ul{a}ttribution and \ul{c}ontribution \ul{t}racking. \sys quantifies each prompt token’s influence on the produced answer, enabling contrasts between pre– and post–unlearning that reveal what changes exactly.

Then, we experiment with six unlearning methods, three LLMs, and three unlearning benchmarks, aiming to answer three questions closely associated to unlearning: 
\begin{packeditemize}
    \item \textbf{Why unlearning can work?} \\
    \textit{It primarily disrupts LLMs' focus on keywords which support the correct answer.}
    
    \smallskip
    \item \textbf{Is knowledge really unlearned?} \\
    \textit{The target knowledge is often not truly erased. Only by explicitly emphasizing relevant keywords in prompt, much of the supposedly ``unlearned'' knowledge can be recovered.}

    \smallskip
    \item \textbf{Why catastrophic forgetting happens?} \\
    \textit{It arises from indiscriminate penalization of all tokens during unlearning, including common words that are essential for general performance.}
    \end{packeditemize}

Taken together, our results articulate an \textit{\textbf{unlearning dilemma}} (shown in \autoref{fig:dilemma}): existing unlearning methods tend to be either recoverable -- keyword emphasis revives the ``forgotten'' knowledge, or overly destructive -- disruption over all prompt tokens induces catastrophic forgetting. A reliable unlearning sits in a narrow and unstable middle ground, challenging the feasibility of achieving it.

\section{Preliminaries}
\label{sec:preliminaries}

\noindent \textbf{Unlearning methods.} We consider the following unlearning methods:
\begin{packeditemize}
\item \textbf{Gradient Ascent} (\GA) \citep{jang2022knowledge} reverses the conventional gradient descent optimization by maximizing  the loss function. This approach represents the most direct unlearning strategy, where the model explicitly increases prediction loss on the data intended to be forgotten.

\item \textbf{Negative Preference Optimization} (\NPO) \citep{zhang2024negative} extends the Direct Preference Optimization (DPO) framework \citep{rafailov2023direct} by treating forgetting data as negative preferences. It reduces the likelihood of generating forgotten content while maintaining proximity to the original model by omitting positive preference terms from the DPO objective.

\item \textbf{Task Vector} (\TV) \citep{ilharco2022editing} manipulates model weights by subtracting the portion related to the forgetting set. It first overtrains the learned model parameterized by $\theta_{l}$ on the forgetting set to obtain the parameters $\theta_{over}$, then computes the task vector $\theta_{over}-\theta_{l}$ and subtracts it from the original model. Then we get unlearned model with parameters $\theta_u = \theta_{l}-(\theta_{over}-\theta_{l})$.

\item \textbf{Representation Misdirection for Unlearning} (\RMU) \citep{li2024wmdp} employs a dual loss mechanism: the forget loss steers activation vectors of hazardous data toward random directions to disrupt internal representations, while the retain loss maintains activation patterns of benign data through regularization to preserve general capabilities.

\end{packeditemize}
Also, we consider the following two regularization loss to maintain performance:
\begin{packeditemize}
\item \textbf{Gradient Descent} (\GDR) \citep{maini2024tofu,zhang2024negative} simply use the prediction loss during training on the retain set, with a standard gradient descent learning objective.
\item \textbf{KL Divergence Minimization} (\KLR) \citep{maini2024tofu,zhang2024negative} minimizes the KL divergence of the prediction distribution between pre- and post-unlearning models. 
\end{packeditemize}
Since \TV and \RMU is incompatible with regularizing \citep{shi2024muse}, we combine \GA and \NPO with \GDR and \KLR, yielding four new combinations. Hence, we end up with a total of six unlearning methods: \GAGDR, \GAKLR, \NPOGDR, \NPOKLR, \TV, and \RMU  (see details in \autoref{app:preliminaries}).

\noindent \textbf{Models.} We choose three LLMs with different architectures and sizes:
\begin{packeditemize}
\item \textbf{Llama-2-7B} \citep{touvron2023llama} is a foundational language model developed by Meta, designed for general-purpose natural language understanding and generation tasks.
\item \textbf{Llama-3.1-8B-Instruct} \citep{grattafiori2024llama3} is Meta's instruction-tuned model optimized for conversational interactions, with enhanced performance on general capabilities.
\item \textbf{Qwen3-14B} \citep{yang2025qwen3technicalreport} is developed by Alibaba Cloud, featuring strong capabilities in various tasks including text generation and question answering with enhanced reasoning abilities.
\end{packeditemize}

\smallskip
\noindent \textbf{Datasets.} We consider three types of textual data in unlearning scenario:
\begin{packeditemize}
\item \textbf{News} \citep{li2023avoiding} consists of BBC news articles collected after August 2023. 
\item \textbf{Books} \citep{eldan2310s} consists of the Harry Potter book series. 
\item \textbf{WMDP} \citep{li2024wmdp} consists of multiple-choice questions covering hazardous biosecurity, cybersecurity and chemistry. For this corpus, we focus on biosecurity domains.
\end{packeditemize}
Notably, unlearning is performed on the text corpus, and evaluation is done on question-answer pairs from the corpus (see details in \autoref{app:preliminaries}). This enables to quantify LLMs' knowledge memorization \citep{shi2024muse}.

\smallskip
\noindent \textbf{LLM-as-a-judge.} Instead of the widely used automatic metric \textsc{Rouge-L} \citep{lin2004rouge}, we employ GPT-4o-mini \citep{hurst2024gpt} as a judge to evaluate whether a model's response matches the ground truth answer \citep{gu2024survey,wei2024systematic}. 
We adopt this approach because we observe that \textsc{Rouge-L} frequently misjudges semantic equivalence (see \autoref{app:rougeL} for details).

\section{\sys: an Interpretable Framework}
\label{sec:unpact}

Prompts directly affect LLM outputs and are accessible in both open- and closed-source settings, different to internal activations or weight states which are only accessible in open-source settings. This motivates us to design our interpretable framework for unlearning from the prompt perspective. Inspired by recent advances in prompt interpretability \citep{cui2025smiley,zhang2024understanding, miglani2023using}, we propose \sys, an interpretable framework for unlearning through prompt attribution and contribution tracking. 

\begin{figure}
    \centering
    \includegraphics[width=0.95\linewidth]{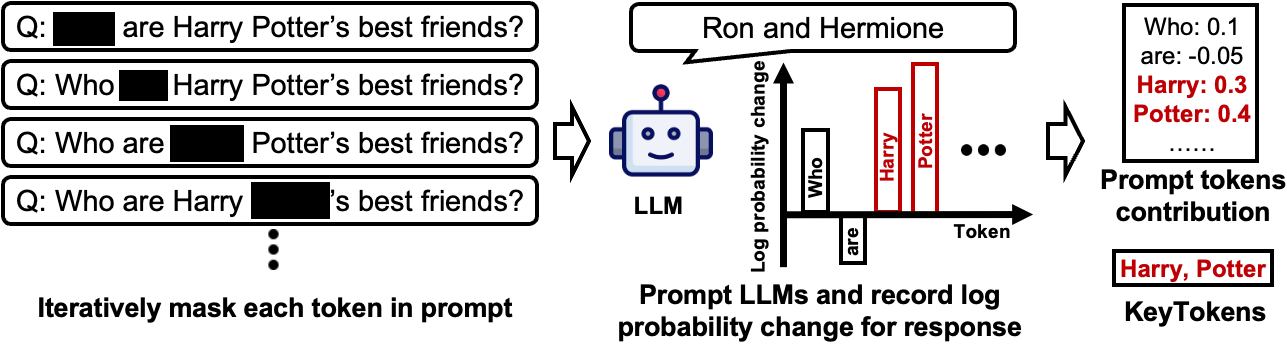}
    \caption{Overview of \sys. It first measures each token's contribution to LLMs' final answers, then gives the \key with the most contributions. 
    Notably, \sys does not require LLMs' internal states and can therefore interpret both open-sourced and closed-sourced models.
    }
    \label{fig:unpact}
    % \vspace{-1em}
\end{figure}

The goal of \sys is to characterize how unlearning reshapes an LLM’s reliance on prompt tokens, thereby revealing what is actually altered by unlearning methods. An overview is shown in \autoref{fig:unpact}. \sys first measures each token's contribution to LLMs' final answers, then gives the \key with the most contributions. 

\subsection{Prompt Tokens Contribution}

The main idea is straightforward: to measure how much a token in the input prompt influences the final output, we perturb the prompt by masking or removing that token and then re-prompt the model. The change in output distribution directly reflects the contribution of the removed token.

Formally, given an input prompt $x=[x_1, x_2, \ldots, x_n]$ and a generated output sequence $y$, the contribution of a token $x_i$ is defined as:
\begin{equation}
    C(x_i,y) = LP(x,y) - LP(x \setminus \{x_i\}, y),
\label{eq:attrapp}
\end{equation}
where $LP(\cdot,\cdot)$ denotes the log probability of generating $y$ conditioned on the input. A positive $C(x_i,y)$ indicates that the token promotes the model’s generation of $y$, while a negative value suggests that the token suppresses it. Thus, $C(\cdot,\cdot)$ quantifies the causal influence of each prompt token on the model’s response.

Since outputs typically contain multiple tokens $y=[y_1, y_2, \ldots, y_T]$, we extend the definition of $LP$ to sequences as:
\begin{equation}
    LP(x,y) = \frac{1}{T} \sum_{k=1}^T LP(x + y_{1:k-1}, y_k),
\label{eq:logProb}
\end{equation}
where $x+y_{1:k-1}$ denotes appending the already generated subsequence $y_{1:k-1}$ to the prompt $x$, separated by a special [SEP] token. This formulation allows us to measure the contribution of prompt tokens across the entire generation trajectory, rather than only the first output token.

Notably, \textit{\textbf{\sys does not rely on internal model states, and can therefore be applied to both open-source and closed-source LLMs.}}

\subsection{\key}

Once the contribution of each prompt token has been computed, we can identify the \key, the subset of tokens that play the most decisive role in determining the model’s output. 
\textbf{\textit{\key reflects the focus of LLMs when generating responses.}}
This intuition is supported by findings in cognitive science, where humans naturally rely on salient words or phrases when processing and recalling information \citep{weingarten2016primed,ellis2016salience}. 
Analogously, in LLM prompting, \key highlights the tokens that the model relies on most heavily to produce the final answer.

Formally, we define \key as the set of tokens with the strongest positive contributions to the answer of LLMs:
\begin{equation}
    K(x,y)=
    \begin{cases}
        \{ x_i \mid C(x_i,y) > 0 \}, & \text{if } |\{x_i \mid C(x_i,y) > 0\}| < \beta \\
        \{ x_i \mid N(C(x_i,y)) > \alpha \}, & \text{otherwise}
    \end{cases}
\end{equation}
where $C(x_i,y)$ denotes the contribution of token $x_i$ for answer $y$, and $N(\cdot)$ normalizes positive contributions across the prompt tokens. 
In practice, the number of tokens with positive contributions can be small. 
To handle such cases robustly, when fewer than a predefined proportion $\beta$ of tokens have positive contributions, we take all of them as \key. 
Otherwise, we apply the normalized threshold $\alpha$ to select only the strongest contributors. 
Both $\alpha$ and $\beta$ are tuned via grid search for stability and generality (see \autoref{app:gridSearch}).

In short, \textit{\textbf{\key is the most influential prompt tokens, serving as a compact lens to reveal how unlearning shifts LLMs' focus within prompt.}}

\section{Why Unlearning Can Work?}
\label{sec:understand}

Since unlearning has been reported as effective in many prior studies \citep{yamashita2025concept,yuan2024closer,ji2024reversing}, we investigate why pre- and post-unlearning models respond differently to questions involving the target knowledge in this section.

\subsection{Experimental Design}

It is worth noting that unlearning is sensitive to hyperparameters \citep{zhong2025dualoptim,kim2024negmerge}. We therefore carefully select configurations that achieve both a high forgetting rate on the forget set and strong utility preservation on the retain set, following the recommendations of \citep{shi2024muse,li2024wmdp} (e.g., training epochs, regularization strength). This ensures that our analysis reflects representative and reasonably optimized unlearning behavior, rather than artifacts of poor parameter tuning. 

% \textbf{Experimental design.} 
For each prompt, we leverage \sys to identify the \key corresponding to the model’s predicted answer. By examining how these influential tokens shift before and after unlearning, we gain an interpretable prompt perspective on what unlearning actually changes inside the model.

\begin{figure*}
    \centering
    \includegraphics[width=0.95\linewidth]{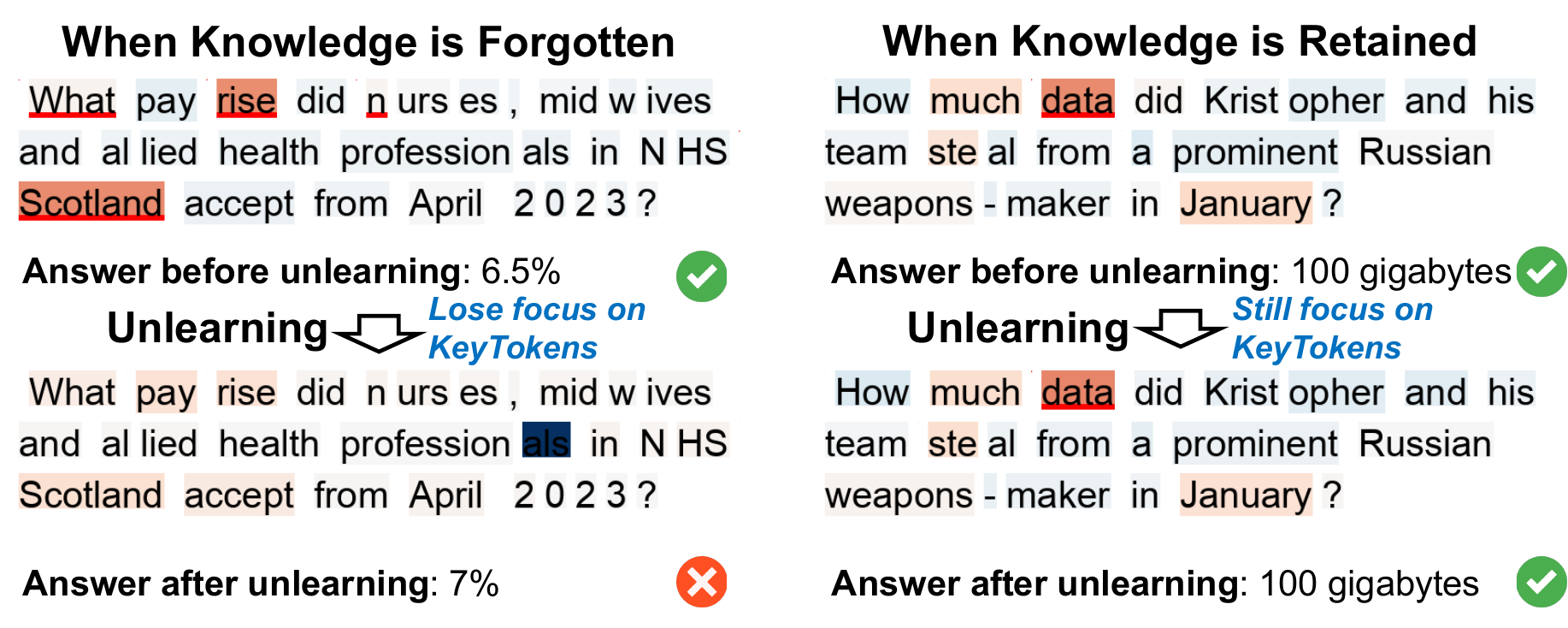}
    \caption{When target knowledge is forgotten, LLMs lose focus on \reduline{\key} ; When target knowledge is retained, LLMs still focus on \reduline{\key} (see more examples in \autoref{app:understand}). \mycolorbox[darkred!60]{Redder} token means more positive contribution; \mycolorbox[darkblue!50]{Bluer} token  means more negative contribution.
    }
    \label{fig:understandUnlearn}
    \vspace{-1em}
\end{figure*}

Specifically, \textbf{\textit{we compare the \key across cases where the target knowledge is either retained or forgotten.}} 
Before unlearning, we apply \sys to the pre-unlearning model to extract the \key $K_{pre}$ associated with the correct answer. 
After unlearning, we apply \sys to the post-unlearning model and distinguish two scenarios: 
$K_{post}^R$, the \key for cases where the model still produces the correct answer (i.e., target knowledge is retained), and 
$K_{post}^F$, the \key for cases where the model produces an incorrect answer (i.e., target knowledge is forgotten). 
By contrasting $K_{pre}$ with $K_{post}^R$ and $K_{post}^F$, we can directly observe how unlearning shifts the model’s focus and thereby reveal what unlearning actually changes inside the model.

\subsection{Results}

% \textbf{Results.} 
\autoref{fig:understandUnlearn} shows that when the target knowledge is forgotten, LLMs lose focus on the \key identified in the pre-unlearning model (e.g., ``What'', ``rise'', ``n'', ``Scotland''). In contrast, when the knowledge is retained, the post-unlearning model continues to attend to the same \key as before (e.g., ``data''). This pattern suggests that \textbf{\textit{unlearning disrupts the association between certain salient keywords and the correct answer}}, mirroring cognitive processes in humans where attention to key information strongly shapes recall and reasoning \citep{tsvilodub2023overinformative,singer1988focused}.

To statistically capture shifts in model focus within the prompt, we adapt cosine similarity (denoted as $CS(\cdot,\cdot)$) \citep{xia2015learning} to compare the \key pre- and post- unlearning. 
Since \key is represented as a set of tokens, we map it into an indicator vector $V(\cdot)$, enabling vector-space comparison. 
We then define a correct focus as:
\begin{equation}
    CS(V(K_{pre}), V(K_{post})) > \gamma,
\end{equation}
where $\gamma$ is a threshold hyperparameter. 
Intuitively, a higher cosine similarity indicates stronger overlap between the \key of the pre-unlearning model ($K_{pre}$) and the post-unlearning model ($K_{post}$), reflecting stability of focus; conversely, lower similarity signals that unlearning has disrupted the association between prompt keywords and the correct answer. 
Implementation details for constructing $V(\cdot)$ are provided in \autoref{app:how2compareKeyTokens}.

\begin{table*}[t]
% \vspace{-8mm}
\centering
% \small
\definecolor{orange}{HTML}{EB8133}

\setlength{\tabcolsep}{8pt}

\renewcommand{\arraystretch}{0.8}
\resizebox{0.95\linewidth}{!}{
\begin{tabular}{cllllllll}
\toprule
  & \GAGDR & \GAKLR & \NPOGDR & \NPOKLR & \TV & \RMU & \AVERAGE\\
 \midrule
\multicolumn{8}{c}{\cellcolor[HTML]{EFEFEF}\textbf{News}} \\
\midrule
Retained & 56.2 & 64.5 & 81.1 & 74.4 & 72.0 & 66.7 & 69.6 \\
Forgotten & 31.9\diffdown{24.3} & 26.7\diffdown{37.8} & 24.0\diffdown{57.1} & 45.5\diffdown{28.9} & 27.3\diffdown{44.7} & 56.9\diffdown{9.8}& 35.4\diffdown{34.2} \\
 \midrule
\multicolumn{8}{c}{\cellcolor[HTML]{EFEFEF}\textbf{Books}} \\
\midrule
Retained & 38.1 & 66.7 & 33.3 & 50.0 & 68.0 & 41.7& 49.6 \\
Forgotten & 31.8\diffdown{6.3} & 27.5\diffdown{39.2} & 23.5\diffdown{9.8} & 33.3\diffdown{16.7} & 27.8\diffdown{40.2} & 25.0\diffdown{16.7}& 28.2\diffdown{21.4} \\
 \midrule
\multicolumn{8}{c}{\cellcolor[HTML]{EFEFEF}\textbf{WMDP}} \\
\midrule
Retained &  14.6& 20.7 & 20.6 & 18.9 & 42.5 & 28.9 & 24.3 \\
Forgotten & 13.9\diffdown{0.7} & 11.5\diffdown{9.2} & 12.0\diffdown{8.6} & 10.9\diffdown{8.0} & 39.8\diffdown{2.7} & 20.0\diffdown{8.9} & 18.0\diffdown{6.3} \\

\bottomrule 
\end{tabular}
}
\caption{The proportion of cases where the model maintains a correct focus on \key pre- and post-unlearning (\%). It drops when the target knowledge is forgotten compared to when it is retained, indicating that unlearning works by shifting the model's focus on \key. }
\label{tab:understand}

\end{table*}

\autoref{tab:understand} reports the proportion of cases where the model maintains a correct focus. 
This proportion drops when the target knowledge is forgotten compared to when it is retained. 
The decline is the most on \normalsize{News} dataset, with an average decrease of $34.2\%$ across different unlearning methods. 
We attribute this to the nature of news content: because the information is relatively recent, the model’s memory of it is likely shallower and more reliant on prompt level. 
As a result, the shift in focus within prompt is clearly revealed by \sys. 
In contrast, the decline is smaller on \textsc{WMDP} dataset. 
Since WMDP consists of multiple-choice questions, the structured format constrains the model’s response space and reduces its reliance on knowledge-related keywords, making the distinction between forgotten and retained knowledge less significant.

Therefore, our results suggest that \textbf{\textit{the apparent effectiveness of unlearning may stem from a disruption of the model’s focus on prompt keywords which support the correct answer.}}

Moreover, since unlearning mainly alters models’ focus within the prompt, the question remains whether the target knowledge is truly erased. We investigate this in the following section.

\section{Is Knowledge Really Unlearned?}
\label{sec:recover}

It is known that knowledge intended to be forgotten can often be restored through interventions such as model weight editing \citep{patil2023can} or additional fine-tuning \citep{hu2025unlearning,lynch2024eight}. 
However, these approaches directly modify the model parameters, effectively producing a new model rather than probing the same post-unlearning one. 
Therefore, a more compelling question is \textbf{\textit{whether forgotten knowledge can be recovered within the same post-unlearning model under a black-box setting. }}

\subsection{Experimental Design}

To investigate this, we impose \textbf{\textit{a stricter condition to recover knowledge: only the input prompt can be modified, and the model is restricted to greedy decoding.}} 
Building on our previous finding in \autoref{sec:understand} that unlearning works by shifting the model's focus away from certain \key correspond to target knowledge, we hypothesize that emphasizing the correct \key in the prompt could resume the focus and thereby restore the supposedly forgotten knowledge.

We therefore design a recovery strategy, \recover, to elicit supposedly ``unlearned'' knowledge from post-unlearning LLMs. 
For each prompt that the pre-unlearning model answers correctly, we first apply \sys to identify the \key $K_{pre}$, the set of keywords on which a correctly answering model places its focus. 
Intuitively, $K_{pre}$ represents the most promising tokens through which the post-unlearning model might recover the forgotten knowledge if they are explicitly emphasized within the prompt. 

To implement this idea, we iterate over subsets $K_{s} \subseteq K_{pre}$ and append to the original prompt a minimal emphasis phrase, such as 
``\textit{Focus on [$K_{s}$] to answer.}'' or 
``\textit{Your answer should focus on [$K_{s}$].}'' 
This simple instruction highlights the relevant keywords without elaborate prompt engineering. 
The effectiveness of such appended phrases is further supported by the recency effect in LLM prompting, where information appearing later in the input often exerts stronger influence \citep{zhang2024understanding,zhao2021calibrate}.

In preliminary experiments, we additionally observed that explicitly including question tokens (e.g., ``How'', ``What'') within prompts further improves recovery (see \autoref{app:questionTokens}). 
A plausible explanation is that models of smaller weight (e.g., 7B, 8B, 14B) exhibit weaker question-answering ability and lower sensitivity to question tokens \citep{han2025question}, which can degrade response quality. 
To mitigate this, we slightly extend $K_{pre}$ by incorporating the question token, ensuring that the model’s attention is guided not only to the knowledge-relevant keywords but also to the question pattern.

\subsection{Results}
\label{sec:recoverRes}

\begin{wrapfigure}{r}{0.47\linewidth}
\vspace{-3ex}
    \centering
    \includegraphics[width=0.99\linewidth]{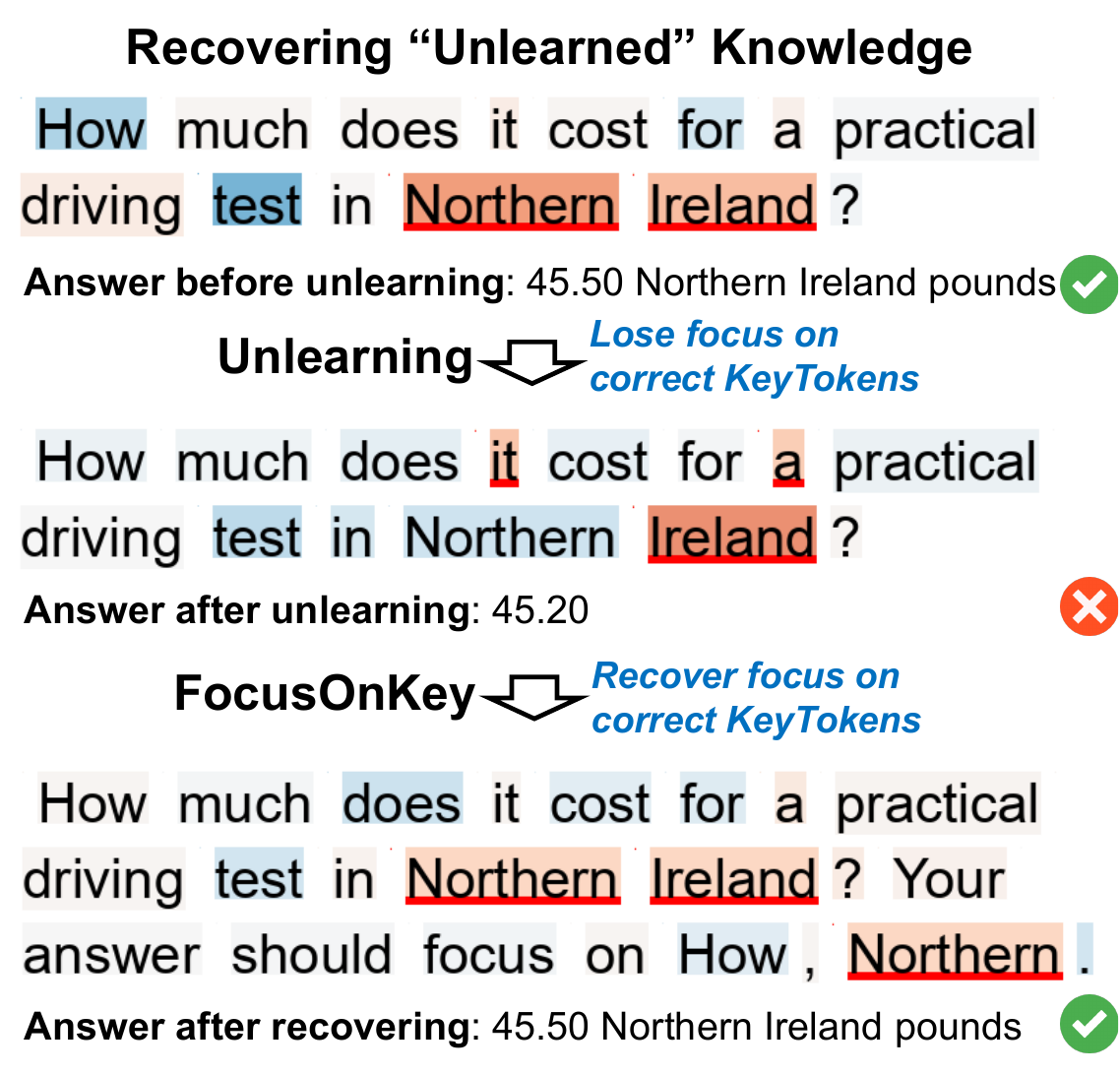}
    \vspace{-2.5ex}
    \caption{Simply emphasizing \reduline{\key} can recover ``unlearned'' knowledge. 
    % \mycolorbox[darkred!60]{Redder} token means more positive contribution; \mycolorbox[darkblue!50]{Bluer} token  means more negative contribution.
    }
    \label{fig:recoverUnlearn}
    \vspace{-3ex}
\end{wrapfigure}

\autoref{fig:recoverUnlearn} illustrates that supposedly ``unlearned'' knowledge can indeed be recovered by \recover. 
For example, the post-unlearning model initially produces an incorrect answer because it ignores the keyword ``Northern.'' 
After appending a simple instruction such as ``Your answer should focus on How, Northern,'' the model re-aligns its attention to ``Northern'' in both the attached phrase and the original question, and consequently generates the correct answer.

\begin{table*}[t]
% \vspace{-8mm}
\centering
% \small
\setlength{\tabcolsep}{8pt}
\definecolor{orange}{HTML}{EB8133}

\renewcommand{\arraystretch}{0.75}
\resizebox{0.95\linewidth}{!}{
\begin{tabular}{cllllllll}
\toprule
 & \GAGDR & \GAKLR & \NPOGDR & \NPOKLR & \TV & \RMU & \AVERAGE\\
 \midrule
\multicolumn{8}{c}{\cellcolor[HTML]{EFEFEF}\textbf{News}} \\
\midrule
\baseline & 12.8 & 25.8 & 53.8 & 39.1 & 66.7 & 4.8 & 33.8 \\
\recover & 27.7\diffup{14.9} & 38.7\diffup{12.9} & 73.1\diffup{19.3} & 69.2\diffup{30.5} & 83.3\diffup{16.6} &9.5\diffup{4.7} & 51.2\diffup{17.4} \\
\midrule
\multicolumn{8}{c}{\cellcolor[HTML]{EFEFEF}\textbf{Books}} \\
\midrule

\baseline & 33.3  & 5.0 & 8.8 & 7.7 & 33.3 & 3.6 & 15.3 \\
\recover  & 42.9\diffup{9.6} & 22.5\diffup{17.5} & 26.5\diffup{17.7} & 28.2\diffup{20.5} & 50.0\diffup{16.7} & 14.3\diffup{10.7}& 30.7\diffup{15.4} \\
\midrule
\multicolumn{8}{c}{\cellcolor[HTML]{EFEFEF}\textbf{WMDP}} \\
\midrule
\baseline  & 33.3& 26.0 & 9.4  &  36.3 & 41.0 & 41.9 & 31.3\\
\recover & 60.3\diffup{27.0} & 49.3\diffup{23.3} & 48.6\diffup{39.2} &39.6\diffup{3.3} & 65.2\diffup{24.2} & 63.4\diffup{21.5} & 54.4\diffup{23.1} \\
\bottomrule
\end{tabular}
}
\caption{Recover rate (\%) of \recover compared to probabilistic evaluation (abbreviated as \baseline) \citep{scholten2024probabilistic}.
\recover (i.e., greedy decoding) recovers approximately 2 times more ``unlearned'' knowledge than \baseline (i.e., multinominal sampling).
}
\label{tab:recover}
\vspace{-1ex}
\end{table*}

To quantify this effect, we define the \emph{recovery rate} as the proportion of supposedly forgotten knowledge that can be restored. 
As shown in \autoref{tab:recover}, \recover achieves strong recovery performance, with rates ranging from $30.7\%$ to $54.4\%$ on average across six unlearning methods. 
We also compare against a baseline \baseline \citep{scholten2024probabilistic}, which approaches recovery from a probabilistic perspective: by sampling multiple outputs rather than relying on greedy decoding, it increases the chance of surfacing forgotten knowledge. 
Nevertheless, \recover attains nearly twice the recovery rate of \baseline while relying solely on greedy decoding.

Notably, this implies that explicitly emphasizing the \key in the prompt elevates forgotten knowledge to the model’s Top-1 prediction, rather than leaving it as a low-probability option recoverable only through multinomial sampling as \baseline. 
In other words, even without exploring multiple trials, \recover substantially outperforms \baseline, underscoring the extent to which unlearned knowledge remains readily accessible through prompt manipulation.

Therefore, our results suggest that \textit{\textbf{the target knowledge is often not truly erased by unlearning methods. 
Only by explicitly emphasizing relevant keywords in prompt, the supposedly ``unlearned'' knowledge can be recovered.}}

\section{Why Catastrophic Forgetting Happens?}
\label{sec:cf}

It is also known that unlearning can induce \emph{catastrophic forgetting} \citep{}, where the model unintentionally loses other useful knowledge unrelated to the target, or even degenerates to producing meaningless outputs (e.g., ``........''). 
Such behavior typically arises when the unlearning strength is too high, for example when no regularization is applied or when training continues for excessive epochs. 
We examine this phenomenon in this section.

Building on our earlier finding in \autoref{sec:understand} that unlearning disrupts models' focus on prompt tokens, we hypothesize that catastrophic forgetting may arise from a similar mechanism. 
To test this, we apply \sys to analyze cases of catastrophic forgetting.

\begin{wrapfigure}{r}{0.46\linewidth}
    \vspace{-2.5ex}
    \centering
    \includegraphics[width=0.99\linewidth]{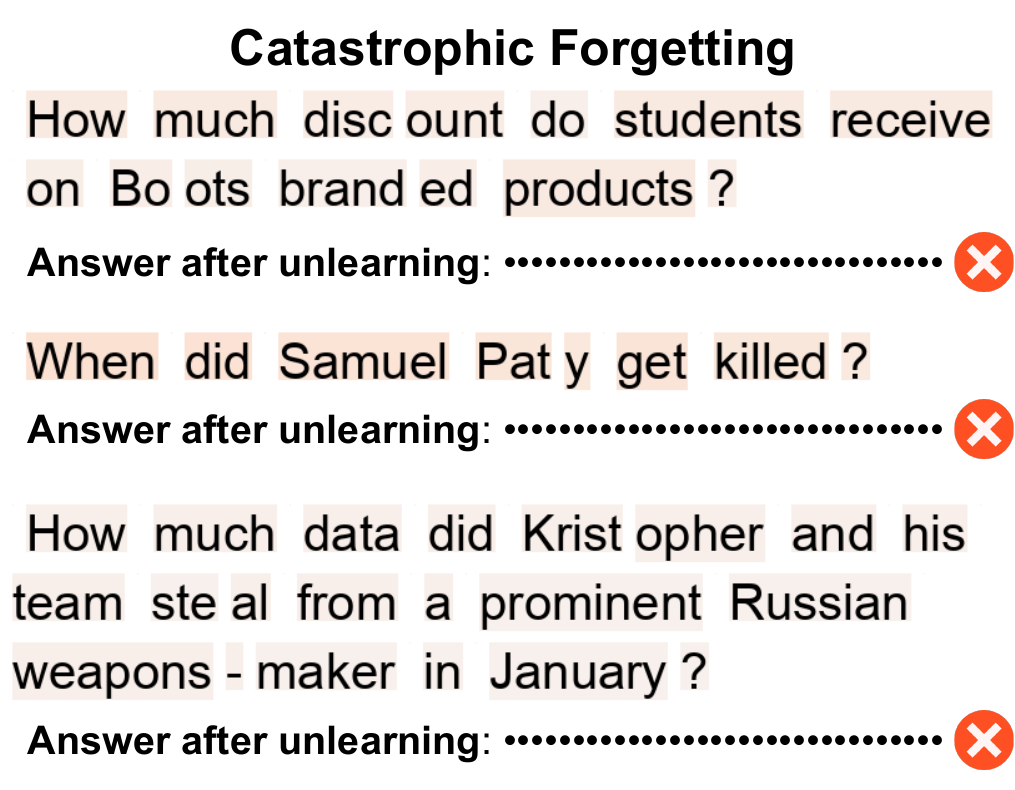}
    \caption{All prompt tokens are ignored when unlearning elicits catastrophic forgetting, where LLMs generate nonsense outputs.}
    \label{fig:catastrophicForgetting}
    \vspace{-1em}
\end{wrapfigure}

\autoref{fig:catastrophicForgetting} shows that when catastrophic forgetting occurs, the LLM neglects \emph{all} tokens in the prompt, and consequently generates nonsensical outputs regardless of the input prompt. 
This interpretation also aligns with the mechanism of unlearning algorithms: they penalize the model on input text containing the target knowledge. 
However, these texts inevitably contain many irrelevant tokens alongside the target ones. 
For example, in the prompt of \autoref{fig:catastrophicForgetting} ``When did Samuel Paty get killed?'', only ``Samuel Paty'' and ``killed'' are directly tied to the target knowledge, while common words such as ``When'', ``did'', and ``get'' are not. 
However, since the unlearning loss does not distinguish between relevant and irrelevant tokens, it inadvertently disrupts the model’s reliance on the latter as well, leading to a collapse in general performance.

In short, our results suggest that \textit{\textbf{catastrophic forgetting arises from indiscriminate penalization of all tokens during unlearning, including common words (e.g., ``get'', ``do'', ``on'') that are essential for LLMs' general performance.}}

\section{The Dilemma of Unlearning}
% \noindent \textbf{Data entanglement?}

Taken together, our findings indicate that current unlearning methods face a dilemma which is hard to avoid: 
either the supposedly ``unlearned'' knowledge remains recoverable, or the model suffers catastrophic forgetting that collapses general performance. 
Yet a reliable unlearning method should achieve both goals simultaneously: making knowledge unrecoverable while preserving normal performance \citep{nguyen2025survey,deep2025unlearning}.

To better understand how far current approaches are from this goal, we introduce a pair of complementary metrics. 
The \emph{recovery rate} (defined in \autoref{sec:recoverRes}) quantifies how easily forgotten knowledge can be restored. 
The \emph{destructive rate} measures the extent to which the model produces irrelevant or nonsensical answers, serving as an indicator of catastrophic forgetting. 
Together, these two metrics capture the trade-off between insufficient unlearning and overly destructive unlearning.

\begin{wrapfigure}{r}{0.5\linewidth}
    \vspace{-3ex}
    \centering    \includegraphics[width=0.99\linewidth]{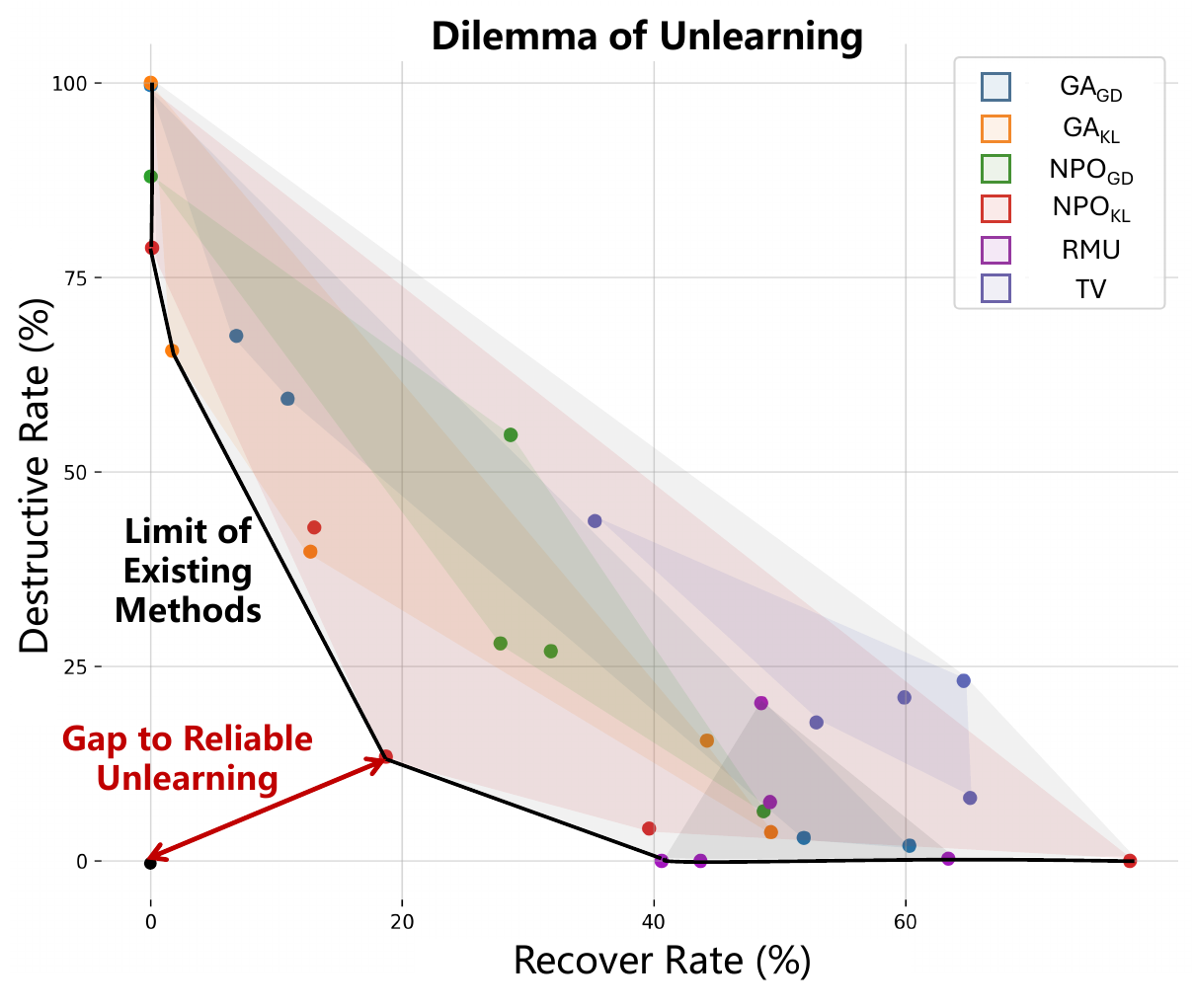}
    \vspace{-2ex}
    \caption{Current unlearning methods face a dilemma, remaining a gap to reliable unlearning.}
    \label{fig:dilemmaRes}
    \vspace{-2.5ex}
\end{wrapfigure}

For each unlearning method, we adopt hyperparameters recommended as optimal in milestone works \citep{shi2024muse,li2024wmdp} to ensure fair comparison. 
To probe catastrophic forgetting, we further increase the training epochs slightly beyond the recommended values, which is known to elicit stronger forgetting effects. 
During the unlearning process, we save checkpoints at every $20\%$ of progress and evaluate both recovery rate and destructive rate across these checkpoints.

\autoref{fig:dilemmaRes} illustrates the performance regions of different unlearning methods, represented by the minimum convex polygons covering all five checkpoints during training. 
Each method exhibits distinct strengths and weaknesses. 
For instance, \NPOKLR evolves from high recoverability toward high destructiveness as training progresses, with an intermediate point that reflects a partial trade-off between the two metrics. 
\RMU is less prone to catastrophic forgetting but also struggles to further reduce the recovery rate. 
To approximate the performance frontier, we connect the checkpoints closest to the origin (which represents reliable unlearning). 
The resulting boundaries highlight that current mainstream unlearning methods still remain a gap to reliable unlearning.

In short, our results reveal that \textbf{\textit{existing unlearning methods trade off between the dilemma of insufficient forgetting and catastrophic forgetting, leaving reliable unlearning still out of reach.}}

\section{Related work}

\noindent \textbf{Machine unlearning.} Generally speaking, machine unlearning seeks to eliminate specific information from a model while preserving its overall performance \citep{thaker2025position,jia2024soul,yuan2024closer,bourtoule2021machine}. 
Early research works primarily address classification tasks \citep{pawelczyk2023context,wang2023kga,tanno2022repairing,guo2019certified}. More recent works have expanded the scope to generation tasks of LLMs \citep{yuan2024closer,li2024wmdp,sheshadri2024latent,maini2024tofu}, which our work focuses on. Mainstream methods of LLMs unlearning primarily rely on parameter optimization, which needs to fine-tune the model on a forgetting set \citep{tamirisa2024tamper,choi2024snap}. However, such methods are likely to harm the overall performance, resulting in catastrophic forgetting \citep{huang2024mitigating,li2024revisiting,liu2024more}. Researchers therefore propose regularizers for utility preservation that either improve the performance on a retain set \citep{liu2022continual} or ensure the unlearned model remains close to the target model during unlearning \citep{maini2024tofu}. 
% As for the target datasets to unlearn,...
% datasets; benchmarks; methods

\smallskip
\noindent \textbf{Questions on unlearning.} Despite the growing interest of unlearning, some recent works question its effectiveness from different perspectives. \citet{ji2024reversing} shows that unlearning suffers from a challenge of catastrophic forgetting, where LLMs' performance degrades significantly \citet{li2024revisiting,luo2023empirical}. \citep{hong2024intrinsic,jin2024rwku,patil2023can} have shown that it is possible to recover unlearned content using existing unlearning heuristics such as residual knowledge, model weights edit. \citet{shumailov2024ununlearning} proposes a concept that when related knowledge is introduced in-context, the unlearned LLMs behave as if it know the forgotten knowledge. \citet{lynch2024eight,hu2025unlearning,qi2023fine} show that fine-tuning on the unlearned data, loosely related data, or even unrelated information can recover unlearned content. However, our work \textit{interprets and recovers unlearned content from a prompt perspective, which is effective for closed-sourced models}.

\smallskip
\noindent \textbf{Interpretability of LLMs.} We summarize the interpretability of LLMs from two aspects. (1) Mechanistic interpretability analyzes model internals to reverse engineer the algorithms learned by the model \citep{belrose2023eliciting,geiger2021causal,elhage2021mathematical,cammarata2021curve}. These works decode intermediate representations that require open-sourced LLMs. 
(2) Prompt interpretability analyzes the input prompt to explain model behaviors. \citep{feng2024unveiling} projects input tokens into the embedding space and estimates their significance. \citep{han2025question} analyzes the impact of specific question tokens on the output. \citep{dong2024promptexp,miglani2023using,zhang2024understanding} leverage prompt tokens' saliency to understand LLMs' behavior. To broaden the use scope to closed-sourced LLMs, we take the latter approach to design \sys, which perturbs each prompt token to see its impact on output logits, without the requirement of LLMs' internal states.

\section{Conclusion}

In this work, we investigate unlearning for LLMs through the lens of interpretability. 
We introduce \sys which quantifies token-level influences and enables direct comparisons between pre- and post-unlearning models. 
Our results reveal that: (1) unlearning appears to be effictive by disrupting LLMs' focus on keywords that support the correct answer; (2) the target knowledge is often not erased and can be recovered through simple keyword emphasis in prompt; (3) catastrophic forgetting arises from indiscriminate penalization of all tokens. 

Taken together, these findings highlight an \textit{unlearning dilemma}: current methods are either insufficient — knowledge remains recoverable, or overly destructive — general performance collapses. 
Reliable unlearning lies in a narrow and unstable middle ground, suggesting that achieving both irrecoverable forgetting and preserved utility remains an open challenge.

\bibliography{main}
\bibliographystyle{iclr2026_conference}

\clearpage
\appendix
\section{Extended Preliminaries}
\label{app:preliminaries}

As mentioned in \autoref{sec:preliminaries}, 
we conducted experiments on Books, News, and WMDP using unlearning methods such as \GA, \NPO, and \TV. These experiments aim to answer three questions closely associated to unlearning. In this section, we will introduce these methods and datasets in detail.

\noindent \textbf{Notations.} Consider an LLM which has learned parameterized by $\theta$, whose hidden states of the model at layer $l$ denoted as $M(\cdot)$. Given an input $x$, it gives the probability distribution over the next tokens $p(\cdot | x; \theta)$. The fine-tuning process on dataset $\mathcal{D} = \{(x_i, y_i)\}_{i=1}^{N}$ aims to minimize the prediction loss $\ell(y|x; \theta) = -\log p(y|x; \theta)$, where $p(y|x; \theta) = \prod_{t=1}^{T} p(y_t | x \circ y_{<t}; \theta)$, $T$ is the number of tokens in the sequence $y$, $y_t$ is the $t$-th token, $y_{<t}$ is the prefix up to $t$, and $\circ$ denotes string concatenation. LLM unlearning requires the unlearned model parameterized by $\theta_u$ to forget a specific forget set $\mathcal{D}_F \subseteq \mathcal{D}$ while maintaining performance on the retain set $\mathcal{D}_R = \mathcal{D} \setminus \mathcal{D}_F$, compared to the initial learned model $\theta_l$.
\subsection{Details of Unlearning Methods}

\begin{packeditemize}
\item \textbf{Gradient Ascent} (\GA) \citep{jang2022knowledge} performs an optimization on the model that is opposite to conventional learning with gradient descent, which is the most straightforward way for unlearning. Specifically, it maximizes the prediction loss on forgetting set.
\begin{equation}
    - \mathbb{E}_{(x,y) \sim \mathcal{D}_F} \left[ -\log p(y \mid x; \theta) \right].
\label{eq:ga}
\end{equation}
\item \textbf{Negative Preference Optimization} (\NPO) \citep{zhang2024negative} builds on DPO \citep{rafailov2023direct} to treat the forget set as negative preference data (ignore positive terms in DPO loss), assigning low likelihood to forgetting set without staying too far from the learned model.
\begin{equation}
- \frac{2}{\mu} \mathbb{E}_{(x,y) \sim \mathcal{D}_R} \left[
\log \sigma \left( -\mu \log \frac{p(y \mid x; \theta)}{p(y \mid x; \theta_{l})} \right)
\right]
\label{eq:npo}
\end{equation}

where $\sigma$ is the sigmoid function and $\mu$ is a hyperparameter which we fix $0.1$ in experiments.
\item \textbf{Task Vector} (\TV) \citep{ilharco2022editing} manipulates model weights to subtract the portion related to the forgetting set. Specifically, we first train the initial learned model $\theta_{l}$ on forgetting set until an overfitting model $\theta_{over}$. We then obtain a task vector related to the weight portion of forgetting by computing the difference $\theta_{over}-\theta_{l}$. Unlearning is achieved by subtracting this task vector from the original learned model $ \theta_u = \theta_{l}-(\theta_{over}-\theta_{l}).$
\begin{equation}
    \theta_u = \theta_{l}-(\theta_{over}-\theta_{l}).
\label{eq:tv}
\end{equation}

\item \textbf{Representation Misdirection for Unlearning} (\RMU)
\citep{li2024wmdp}
Specifically, RMU employs a dual loss function mechanism: the forget loss disrupts the integrity of internal representations by steering the activation vectors of hazardous data toward random directions, rendering the model unable to correctly decode the relevant information; the retain loss ensures that the activation patterns of benign data remain consistent with the original model through regularization, thereby maintaining the model's general capabilities.
\begin{equation}
\mathbb{E}_{x_F \sim \mathcal{D}_F} \frac{1}{T_F}\sum_{x_{F_t} \in x_F}{||M_{u}(x_{F_t})-c \cdot \textbf{u}||_2^2
}+ \alpha \cdot \mathbb{E}_{x_R\sim \mathcal{D}_R} \frac{1}{T_R}\sum_{x_{R_t} \in x_R}{||M_{u}(x_{R_t})-M_{fro}(x_{R_t})||_2^2
}
\label{eq:rmu}
\end{equation}

\end{packeditemize}

We also consider the following two regularization loss to maintain performance:
\begin{packeditemize}
\item \textbf{Gradient Descent} (\GDR) \citep{maini2024tofu,zhang2024negative} simply use the prediction loss during training on the retain set, with a standard gradient descent learning objective. This approach is straightforward and computationally efficient, as it only requires forward and backward passes on the data to be retained. The formula is as follows.
\begin{equation}
    \mathbb{E}_{(x,y) \sim \mathcal{D}_R} \left[ -\log p(y \mid x; \theta) \right].
\label{eq:gd}
\end{equation}
\item \textbf{KL Divergence Minimization} (\KLR) \citep{maini2024tofu,zhang2024negative} minimizes the KL divergence of the prediction distribution between pre- and post-unlearning models. By maintaining distributional consistency, KLR helps prevent catastrophic forgetting and preserves model utility on retained data. The formula is as follows.
\begin{equation} 
\mathbb{E}_{(x,y) \sim \mathcal{D}_R} \left[
\mathrm{KL}(p(y \mid x; \theta) \,\|\, p(y \mid x; \theta_{l}))
\right].
\label{eq:kl}
\end{equation}
\end{packeditemize}
\subsection{Details of Datasets}
We use the models which have learned related knowledge parameterized by $\theta_l$ as the target for unlearning. Notably, unlearning is performed on the text corpus, and evaluation is done on question-answer pairs derived from the corpus, which can quantify LLMs’ memorization of knowledge on $D_F$ and $D_R$. The text and pair are shown in \autoref{tab:datasets}.
\begin{table}[htbp]
\centering
\renewcommand{\arraystretch}{1} % 调整行间距
\begin{tabular}{@{}p{1cm}p{12cm}} % 使用 p{宽度} 来控制列宽
\hline
\textbf{\small{Dataset}} &  \textbf{\small{Example}} \\
\hline
\multirow{3}{*}{\normalsize{News}}& \small{\textbf{Text:} Sweden's Loreen won the competition for the second time with her pop anthem Tattoo.} \\
\cmidrule(lr){2-2}
& \small{\textbf{Evaluated question}: Which song did Sweden's Loreen win the Eurovision Song Contest with?} \newline \small{\textbf{Ground truth answer}: Tattoo} \\
\hline
\vspace{-1mm}
\multirow{3}{*}{\centering\normalsize{Books}} & \small{\centering\textbf{Text:} ``Look there,'' said Hagrid, ``see ... on the ground? ... That's unicorn blood.''} \\
\cmidrule(lr){2-2}
 & \small{\textbf{Evaluated question}: What specific type of blood did Hagrid point out on the ground?} \newline 
\small{\textbf{Ground truth answer}: unicorn blood}\\
\hline
\multirow{4}{*}{\centering\normalsize{WMDP}} & \small{\textbf{Evaluated question}: Which secretion system has been found in B. mallei?\newline A. T2SS, \ \ \ \ \ \ B. T6SS\newline C. Both, \ \ \ \ \ \ D. Neither} \newline\small{\textbf{Ground truth answer}: A} \\ 
\hline
\end{tabular}
\caption{Examples of datasets to unlearn. Unlearning is performed on the text and evaluation is on question-answer pairs derived from the text.}
\label{tab:datasets}
\end{table}

\section{Grid Search of Threshold}
\label{app:gridSearch}

\begin{figure}[h!]
    \centering
    \includegraphics[width=0.6\linewidth]{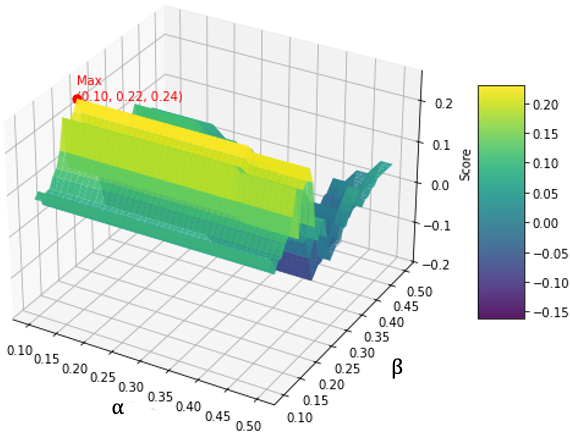}
    \caption{Grid search of hyperparameters $\alpha=0.22, \beta=0.24$.}
    \label{appfig:grid}
    % \vspace{-1.5em}
\end{figure}

As mentioned in \autoref{sec:unpact}, we define \key as follows:
\begin{equation}
    K(x,y)=
    \begin{cases}
        \{ x_i \mid C(x_i,y) > 0 \}, \text{if less than }\beta \\
        \{ x_i \mid N(C(x_i,y)) > \alpha \}, \text{otherwise}
    \end{cases}
\end{equation}
For the hyperparameter $\alpha, \beta$ from this equation, we perform a grid search to determine their values. Specifically, we iterate within the range of $[0.1,0.5]$ for $\alpha, \beta$, and compute the difference of number of focused \key between correct and incorrect response cases. \autoref{appfig:grid} shows that we can determine the best hyperparamters as $\alpha=0.22, \beta=0.24$

\clearpage
\section{Limitation of \textsc{Rouge-L}}
\label{app:rougeL}

Since \textsc{Rouge-L} was proposed by \citep{lin2004rouge}, it has become a widely evaluation metric~\citep{shi2024muse}. It is computed based on the longest common subsequence match, The formula is as follows:
\begin{equation}
R_{lcs} = \frac{\text{LCS}(X,Y)}{m}
\end{equation}
\begin{equation}
R_{lcs} = \frac{\text{LCS}(X,Y)}{n}
\end{equation}
\begin{equation}
\text{\textsc{Rouge-L}} = \frac{(1+\beta^2)R_{lcs}P_{lcs}}{R_{lcs}+\beta^2P_{lcs}}
\end{equation}
where $\text{LCS}(a,b)$ is the length of longest common subsequence between $a$ and $b$, $X$ is a reference sentence and $Y$ is a candidate sentence, whose sizes are respectively $m$ and $n$, $\beta$ is a hyperparameter. 

\textsc{Rouge-L} measures the similarity between two sentences from the perspective of words and phrases. However, it lacks the capability to assess semantic-level equivalence, such as recognizing synonyms. Our experimental findings further corroborate that \textsc{Rouge-L} is not invariably a reliable metric. In certain instances, it fails to adequately capture the similarity to the ground truth.

In the question ``How much data did Kristopher and his team steal from a prominent Russian weapons-maker in January?'', ground truth is ``100gigabytes'', and the LLM responds ``100GB''. Though they are the same in semantic-level assessments, the \textsc{Rouge-L} score is 0. Furthermore, our statistical analysis reveals that 37.5\% of the correct answers possess a \textsc{Rouge-L} score of less than 20. so relying solely on \textsc{Rouge-L} for answer evaluation may lead to misjudgments, especially when there is diversity in the expression of answers.

To more effectively capture semantic-level information and mitigate the impact of synonyms on \textsc{Rouge-L} scores, we employed an LLM during the judge phase. This was achieved by utilizing a prompt template to provide the LLM with the question, ground truth, and the LLM-generated answer. The LLM assesses the correctness and provides ``Yes'' or ``No'' and a rationale for its judgment. The prompt template is presented below.
\begin{tcolorbox}
    [width=\linewidth,colback={white},title={\fontsize{9.5}{7}\selectfont Prompt Template of LLM Judge},coltitle=white,left=1pt,right=1pt,top=1pt,bottom=1pt] 
{\small
You are a grading assistant. Your task is to compare a student's answer with the reference answer and determine whether the student's answer is content-wise equivalent to the reference answer, even if it is expressed in a different way.\\
Please focus on semantic similarity rather than wording or structure. If the student's answer conveys the same key points, facts, and reasoning as the reference answer, it should be considered correct. \\
The student's answer may contain extra information that is unrelated to the question. Please ignore such irrelevant parts.\\
Reference Answer:\\
\placeholder{\{reference\}}\\
\\
Student Answer:\\
\placeholder{\{student\}}\\
\\
Question:\\
\placeholder{\{question\}}\\
\\
Does the student's answer match the reference answer in terms of meaning? Answer ``Yes'' or ``No'', and briefly explain your reasoning.
}
\end{tcolorbox}

\clearpage

\section{How to Compare \key?}
\label{app:how2compareKeyTokens}

As mentioned in \autoref{sec:understand}, we adapt cosine similarity (i.e.,  $CS(\cdot,\cdot)$) \citep{xia2015learning} to quantify the difference between \key before and after unlearning:
\begin{equation}
    CS(V(K_{pre}),V(K_{post})) > \gamma,
\end{equation}
where $V(\cdot)$ transfers a set of tokens to an indicator vector which facilitate computation, and $\gamma = 0.5$ is determined through grid search of hyperparameter.

Specifically, suppose the \key before and after unlearning is (``Harry'', ``Potter'') and (``Harry''). $V(\cdot)$ gives the occurrence list of both \key before and after unlearning, i.e., $V(\text{``Harry'', ``Potter''})=[1,1]$ since ``Harry'':1 and ``Potter'':1, $V(\text{``Harry''})=[1,0]$ since ``Harry'':1 and ``Potter'':0. Then, we compute the cosine similarity $CS([1,1],[1,0])=0.71>0.5$, indicating that the \key before and after unlearning is similar.

% \clearpage
\section{Question Tokens Facilitate Recovery}
\label{app:questionTokens}

\begin{figure}[h]
    \centering
    \includegraphics[width=0.5\linewidth]{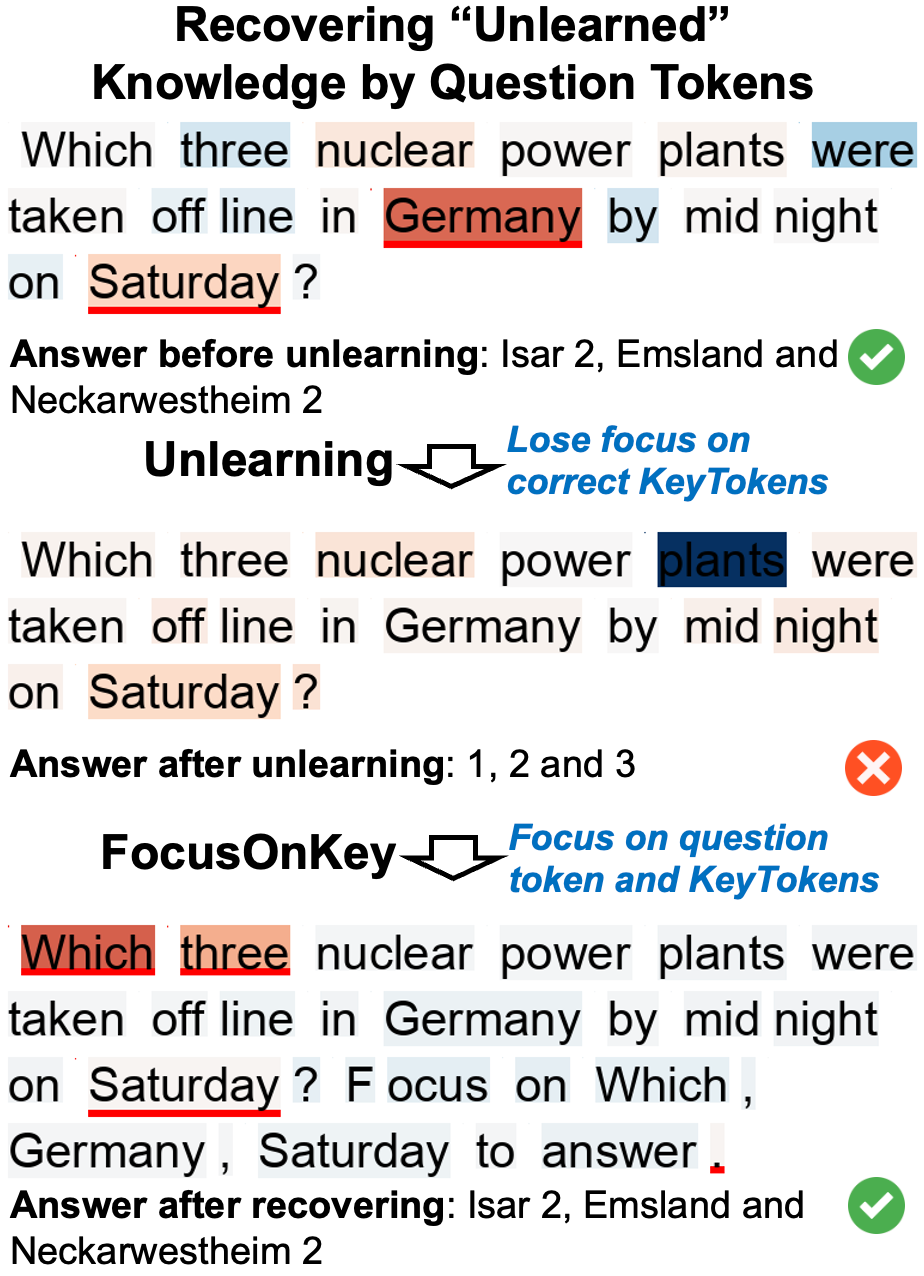}
    \caption{Question token ``Which'' facilitates \key focus and knowledge recovery.
    }
    \label{appfig:questionTokes}
\end{figure}

As mentioned in \autoref{sec:recover}, we observe that question tokens within prompts facilitate knowledge recovery. \autoref{appfig:questionTokes} shows an example. \key for correct response is ``Germany'' and ``Saturday''. Emphasizing the question token ``Which'' along with \key make the LLM focus on the question token itself, and successfully recover focus on ``Saturday''. And the ``unlearned'' knowledge is also recovered.

A plausible explanation of this phenomenon is that the limited question-answering capability of the Llama2-7B model results in lower sensitivity to question tokens \citep{han2025question}, thereby affecting response quality. Therefore, we slightly modify $K_{pre}$ by adding the question token.

\clearpage

\section{Recovering ``Unlearned'' Knowledge}
\label{app:recover}
\begin{figure}[h]
    \centering
    \includegraphics[width=0.99\linewidth]{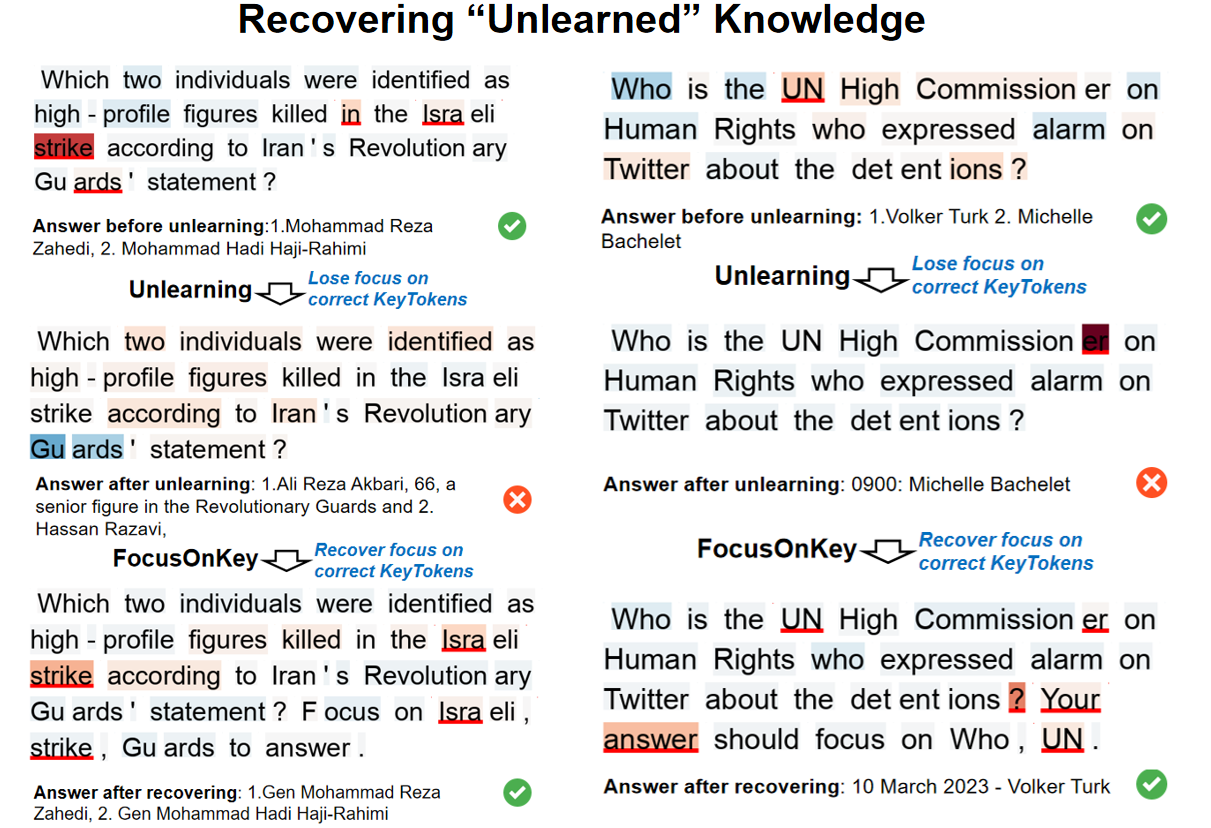}
    \caption{Simply emphasizing \reduline{\key} can recover ``unlearned'' knowledge. \mycolorbox[darkred!60]{Redder} token means more positive contribution; \mycolorbox[darkblue!50]{Bluer} token  means more negative contribution.}
    \label{fig:recover}
\end{figure}

\autoref{fig:recover} shows more examples of recovering ``unlearned'' knowledge unveiled by \sys. Similarly, redder token means more positive contribution; bluer token means more negative contribution. In \autoref{sec:recover}, \autoref{fig:recoverUnlearn} shows “unlearned” knowledge can be readily recovered by focusing on correct \key within prompt.

Following the observation, we turn our attention to the top case for further analysis. Initially, learned LLMs focused on the \key``in'', ``Isra'',``strike'' and ``ards'', therefore answered correctly. After unlearning, none of the \key was focused on, so unlearning works. Then, we applied \recover and discovered that some of \key before unlearning received renewed attention. As a result, the ``unlearned'' knowledge was recovered. 

The analysis of the bottom case aligns above. For initial response, LLMs concentrated on the correct \key, resulting in accurate responses. After unlearning, LLMs shifted their focus to incorrect \key, which led to wrong answers. Subsequently, the application of \recover enabled the LLMs to appropriately refocus on \key in the input question, thereby successfully recovering the previously ``unlearned'' knowledge.

\section{Why Unlearning Can Work?}
\label{app:understand}

\begin{figure}[h]
    \centering
    \begin{subfigure}{\textwidth}
        \centering
        \includegraphics[width=\linewidth]{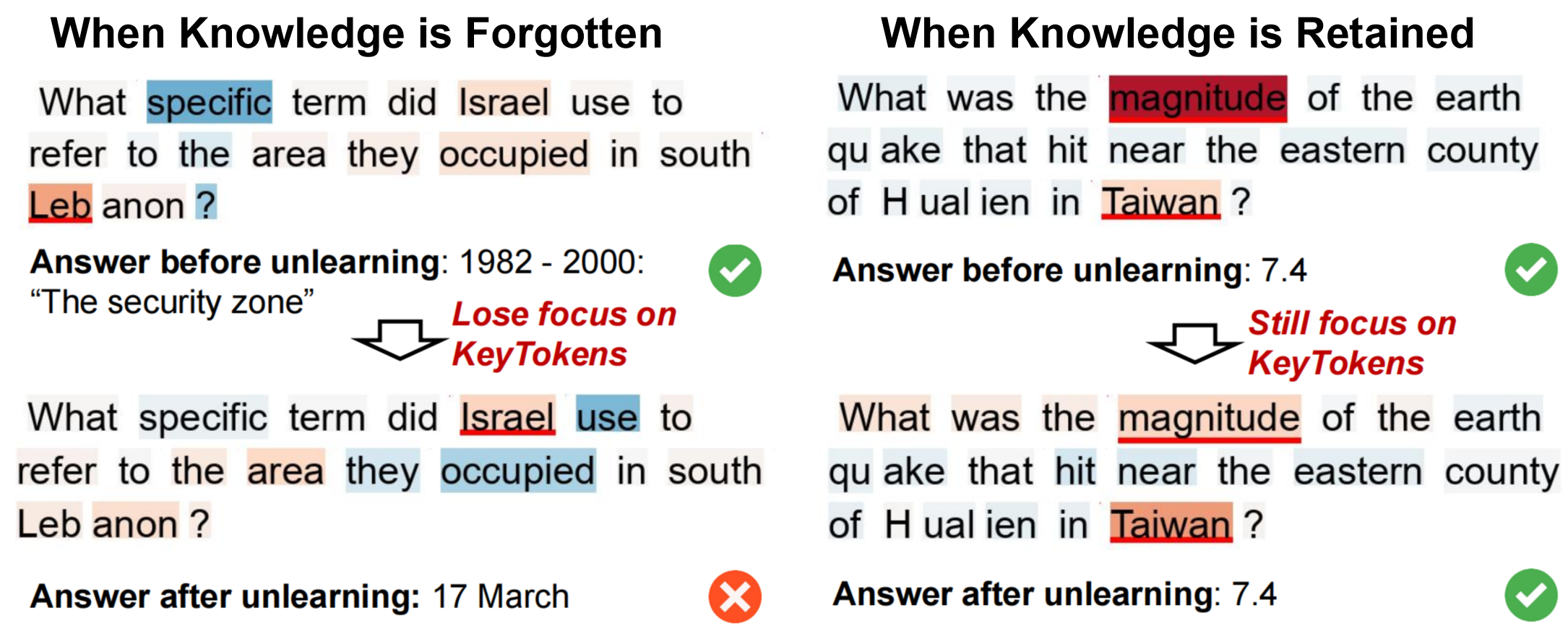}
    \end{subfigure}
    \begin{subfigure}{\textwidth}
        \centering
        \includegraphics[width=\linewidth]{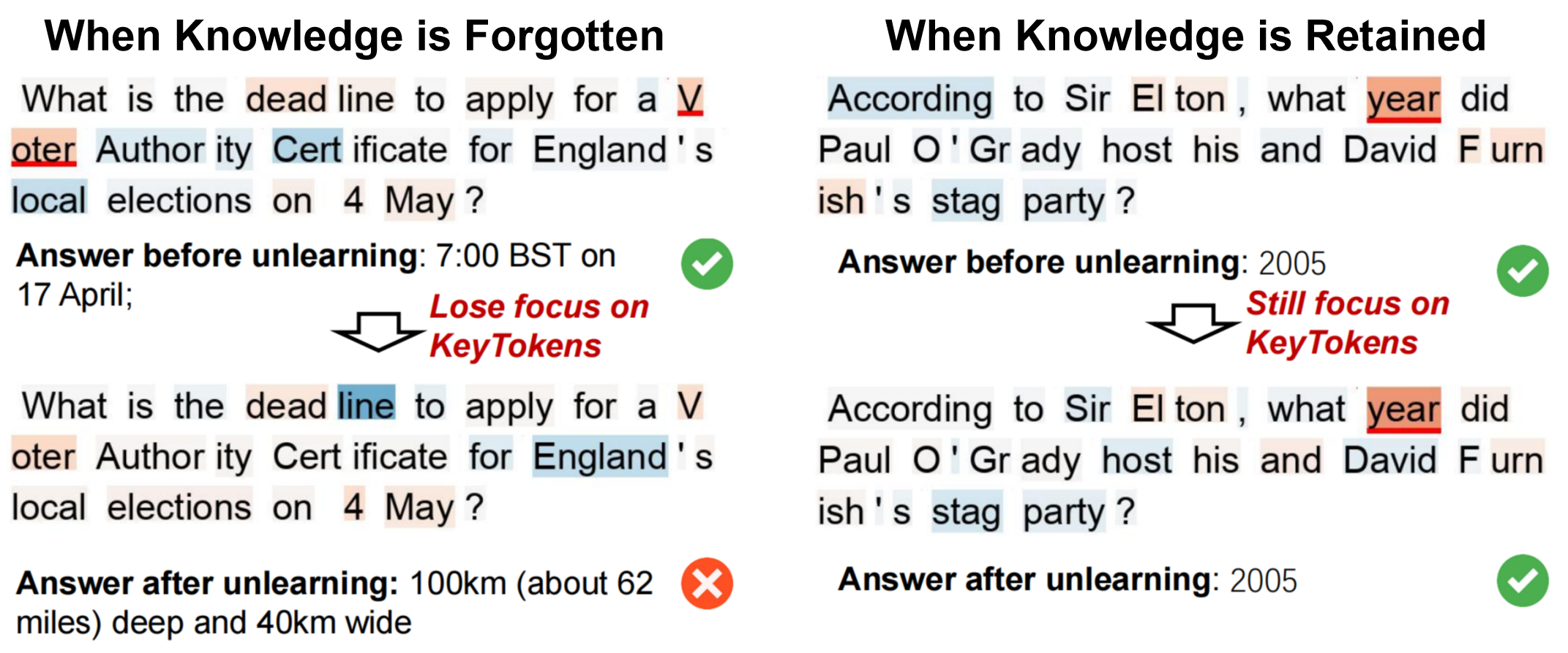}
    \end{subfigure}
    \begin{subfigure}{\textwidth}
        \centering
        \includegraphics[width=0.95\linewidth]{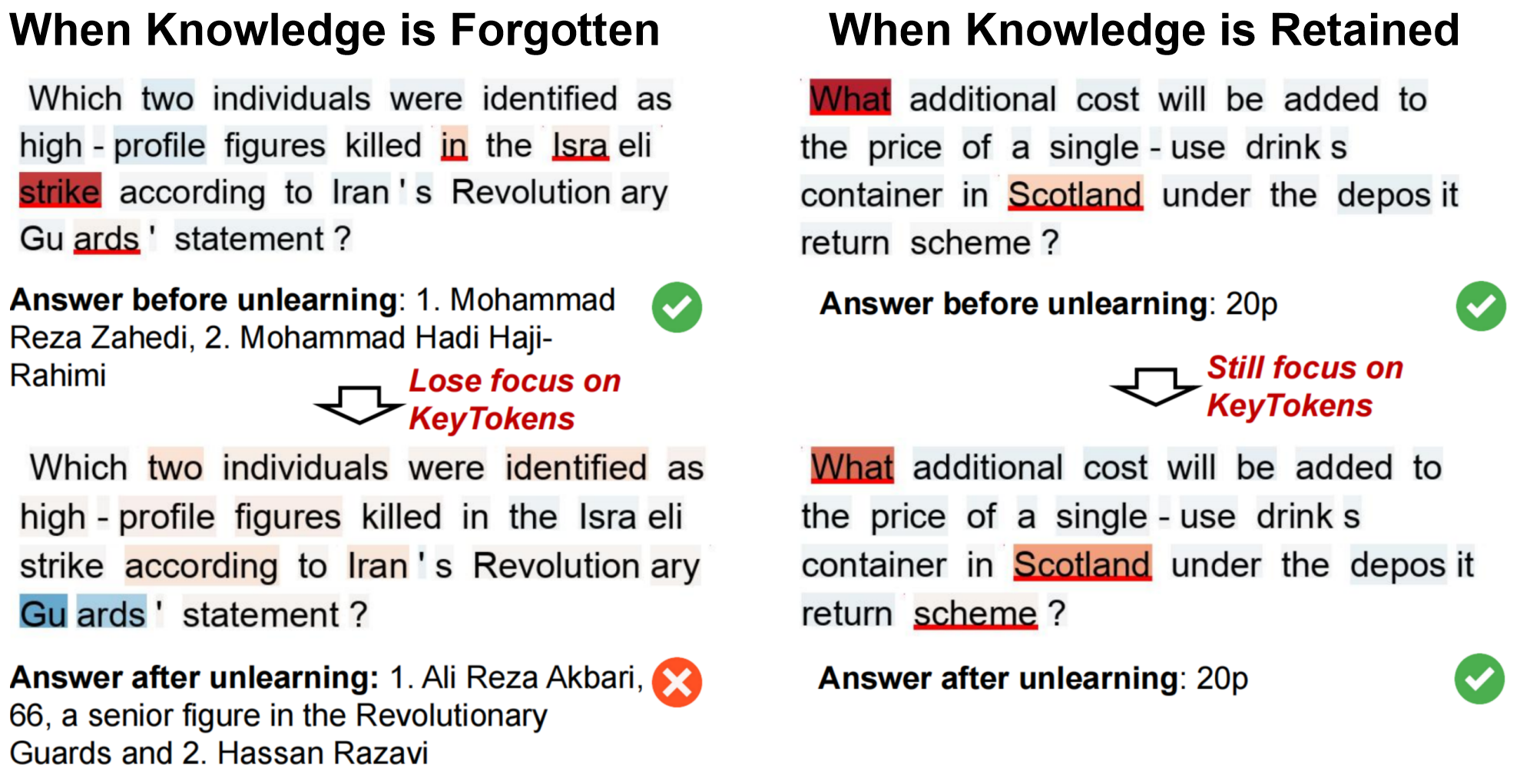}
    \end{subfigure}
    % \vspace{-30pt}
    \caption{When target knowledge is forgotten, LLMs lose focus on \reduline{\key} ; When target knowledge is retained, LLMs still focus on \reduline{\key}. \mycolorbox[darkred!60]{Redder} token means more positive contribution; \mycolorbox[darkblue!50]{Bluer} token  means more negative contribution. This figure is continued in \autoref{fig:WorkFails2}.}
    \label{fig:WorkFails1}
\end{figure}

\begin{figure}[h]
    \centering
    \begin{subfigure}{\textwidth}
        \centering
        \includegraphics[width=\linewidth]{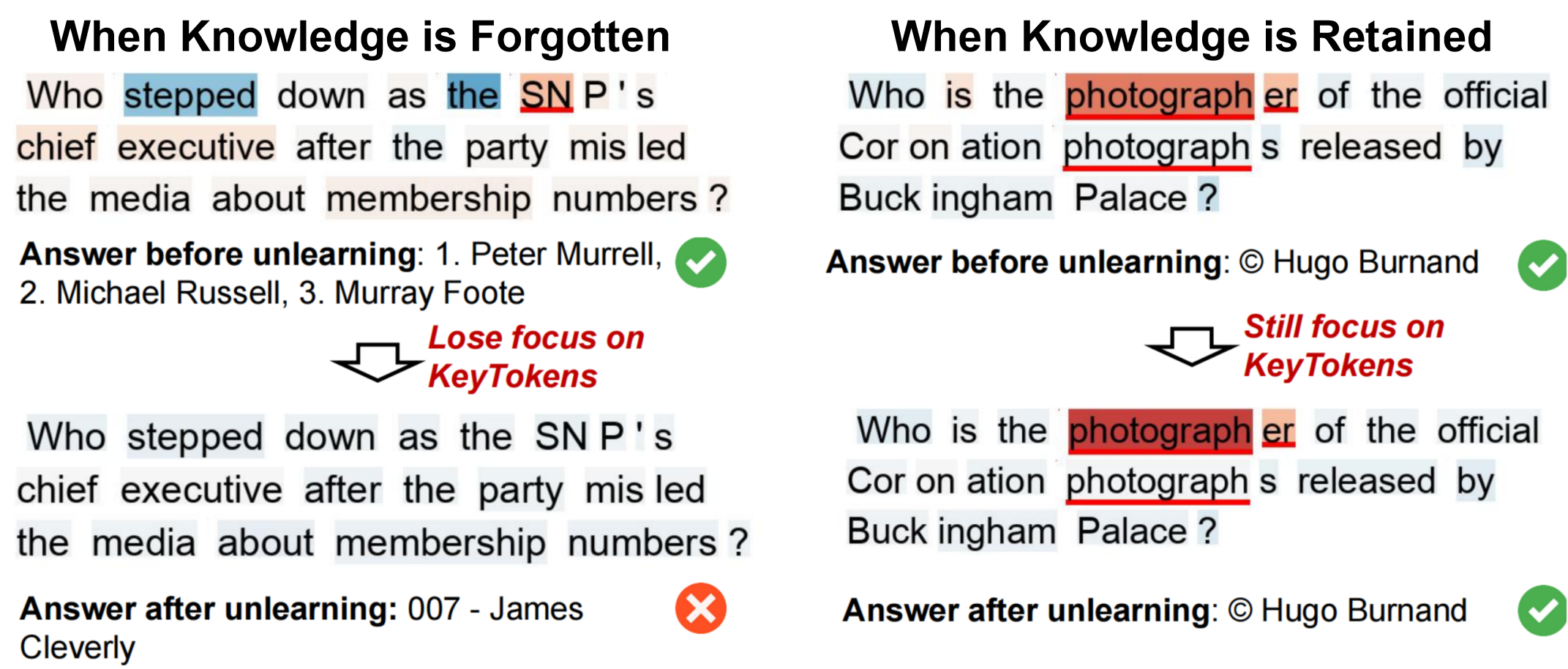}
    \end{subfigure}
    \begin{subfigure}{\textwidth}
        \centering
        \includegraphics[width=\linewidth]{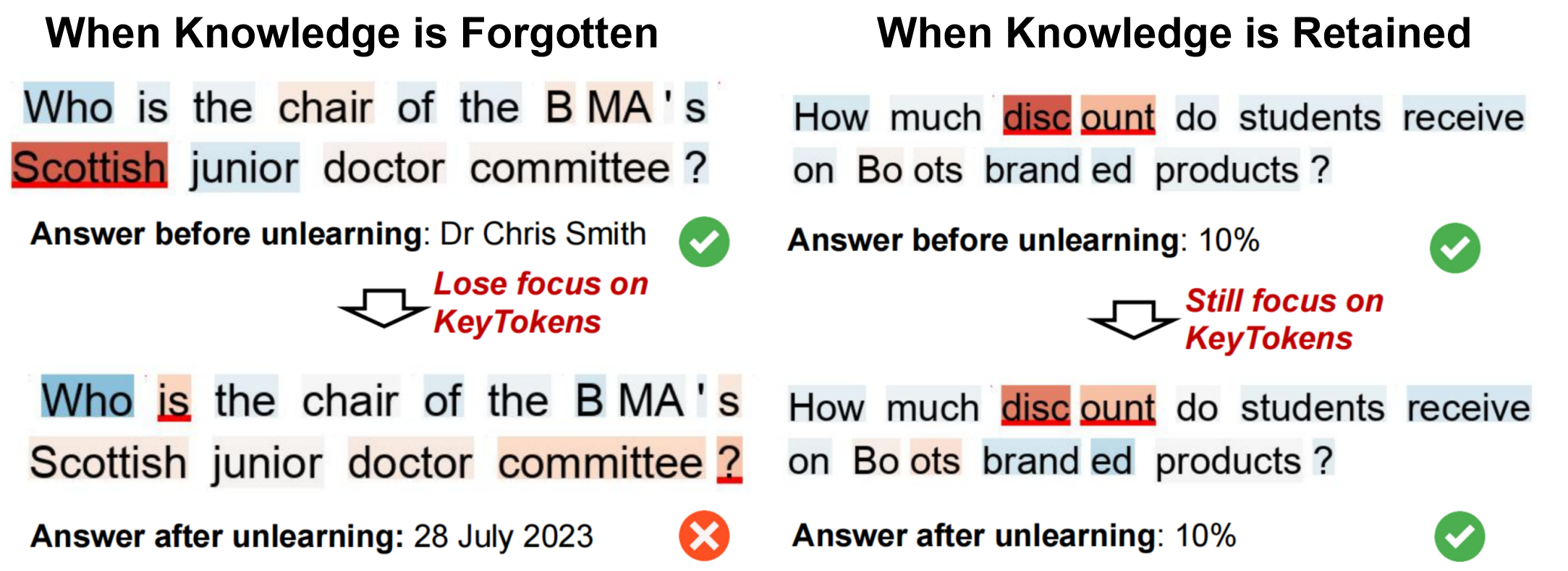}
    \end{subfigure}
    \begin{subfigure}{\textwidth}
        \centering
        \includegraphics[width=\linewidth]{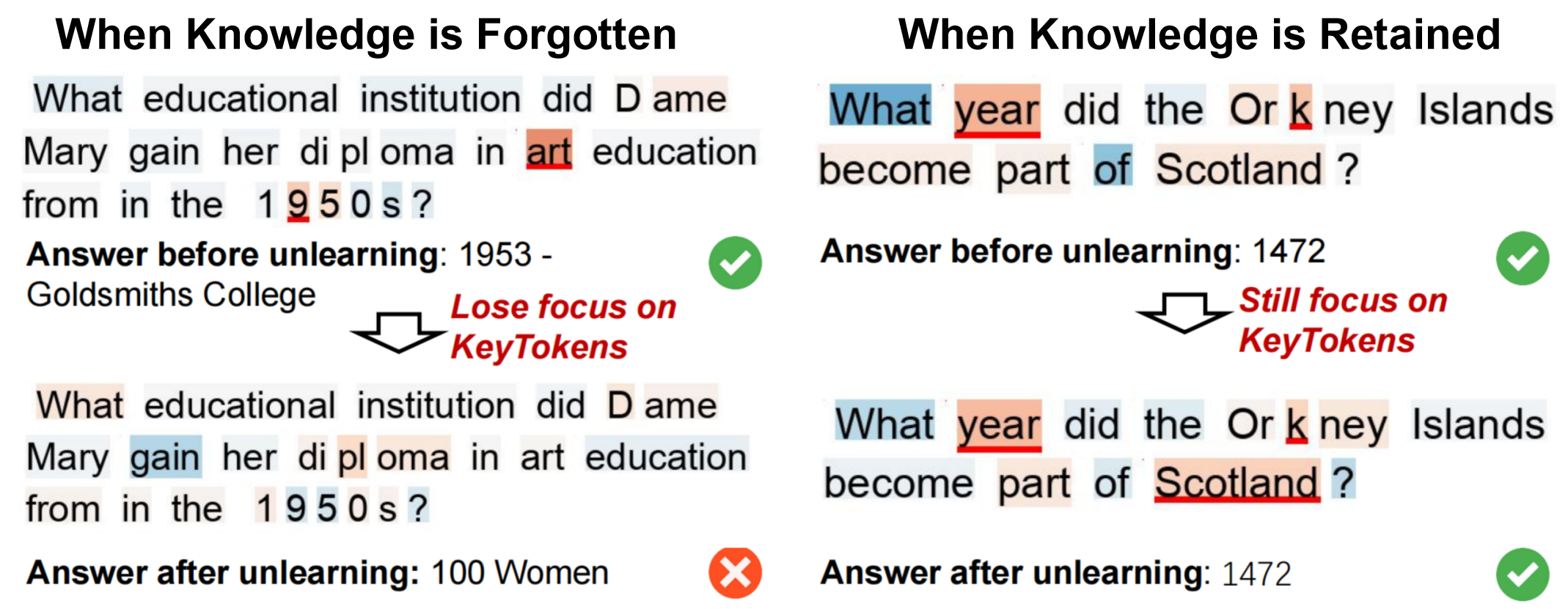}
    \end{subfigure}
    \vspace{-15pt}
    \caption{When target knowledge is forgotten, LLMs lose focus on \reduline{\key} ; When target knowledge is retained, LLMs still focus on \reduline{\key}. \mycolorbox[darkred!60]{Redder} token means more positive contribution; \mycolorbox[darkblue!50]{Bluer} token  means more negative contribution. This Figure is continued from \autoref{fig:WorkFails1} and continued in \autoref{fig:WorkFails3}.}
    \label{fig:WorkFails2}
\end{figure}

\begin{figure}[h]
    \centering
    \begin{subfigure}{\textwidth}
        \centering
        \includegraphics[width=\linewidth]{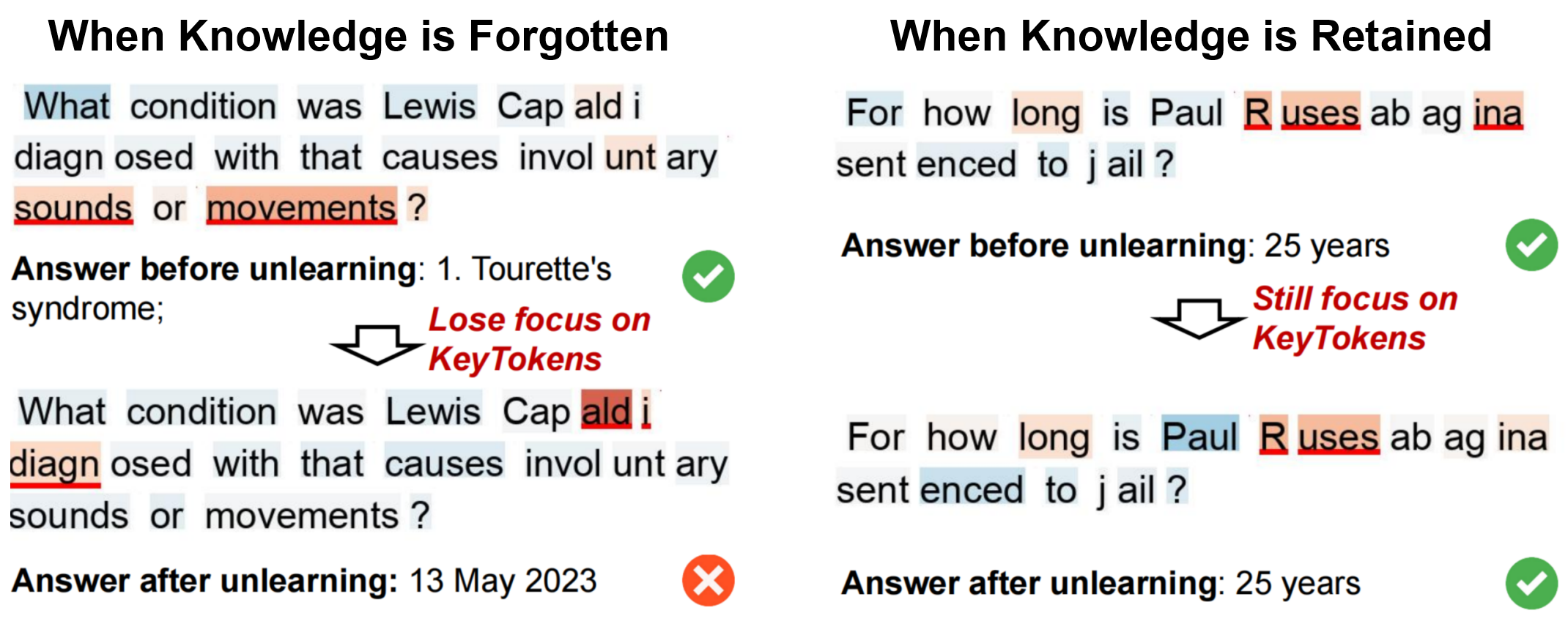}
    \end{subfigure}
    \begin{subfigure}{\textwidth}
        \centering
        \includegraphics[width=\linewidth]{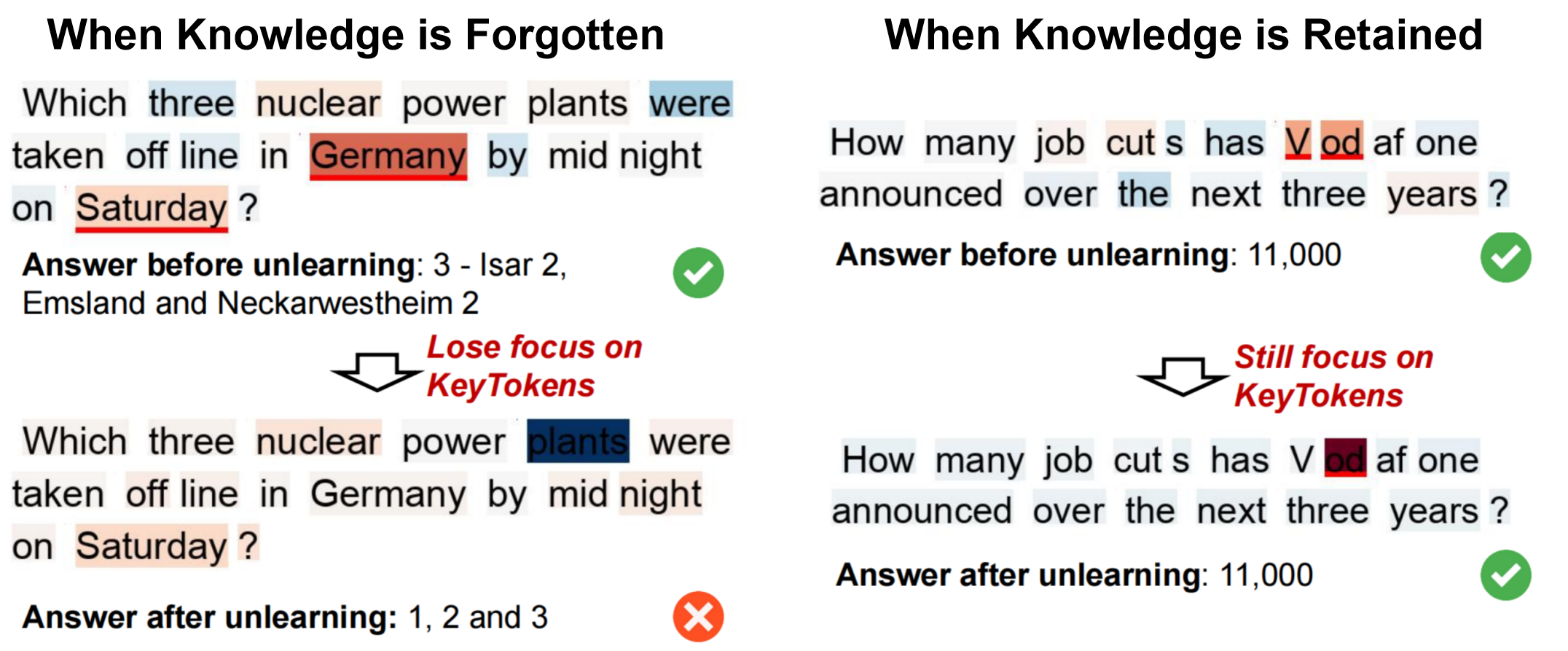}
    \end{subfigure}
    \begin{subfigure}{\textwidth}
        \centering
        \includegraphics[width=\linewidth]{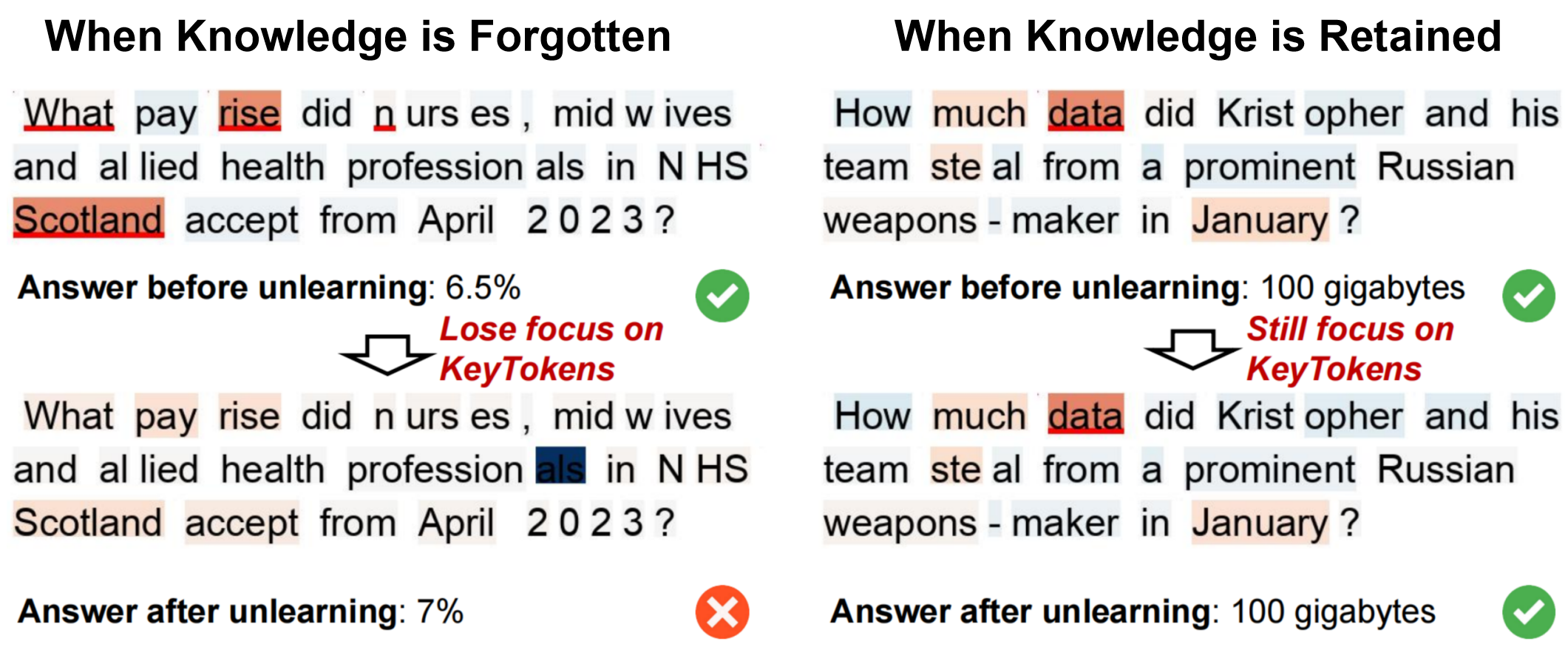}
    \end{subfigure}
    \vspace{-15pt}
    \caption{When target knowledge is forgotten, LLMs lose focus on \reduline{\key} ; When target knowledge is retained, LLMs still focus on \reduline{\key}. \mycolorbox[darkred!60]{Redder} token means more positive contribution; \mycolorbox[darkblue!50]{Bluer} token  means more negative contribution. This figure is continued from \autoref{fig:WorkFails2}.}
    \label{fig:WorkFails3}
\end{figure}

\autoref{fig:WorkFails1}, \autoref{fig:WorkFails2}, and \autoref{fig:WorkFails3} show more examples of \autoref{fig:understandUnlearn} in \autoref{sec:understand}. Redder token means more positive contribution; Bluer token means more negative contribution. In \autoref{sec:understand}, we observed that unlearning succeeds or fails because it disrupts or does not disrupts LLMs’ focus on \key for correct response.


\end{document}